\documentclass[lettersize,journal]{IEEEtran}
\usepackage{amsmath,amsfonts}
\usepackage{algorithmic}
\usepackage{algorithm}
\usepackage{physics}
\usepackage{hyperref}
\hypersetup{colorlinks=true}

\usepackage[capitalize]{cleveref}
\usepackage[caption=false,font=normalsize,labelfont=sf,textfont=sf]{subfig}
\usepackage{textcomp}
\usepackage{stfloats}
\usepackage{url}
\usepackage{verbatim}
\usepackage{booktabs} 
\usepackage{graphicx}
\usepackage{wrapfig}
\usepackage{xcolor}
\usepackage{caption}
\usepackage{cite}
\newcommand{\llava}{LLaVA}
\newcommand{\llama}{LLaMA}
\newcommand{\nerf}{NeRF}
\newcommand{\nftovec}{\texttt{nf2vec}}
\newcommand{\inrtovec}{\texttt{inr2vec}}

\newcommand{\model}{LLaNA}

\begin{document}

\title{Scaling LLaNA: Advancing NeRF-Language Understanding Through Large-Scale Training}

%\author{IEEE Publication Technology,~\IEEEmembership{Staff,~IEEE,}
        % <-this % stops a space
%\thanks{This paper was produced by the IEEE Publication Technology Group. They are in Piscataway, NJ.}% <-this % stops a space
%\thanks{Manuscript received April 199999999999999+96666666669, 2021; revised August 16, 2021.}}

%{Shell \MakeLowercase{\textit{et al.}}: A Sample Article Using IEEEtran.cls %for IEEE Journals}


\author{Andrea Amaduzzi, Pierluigi Zama Ramirez, Giuseppe Lisanti, Samuele Salti, Luigi Di Stefano
\\
CVLAB, University of Bologna
\\
\url{https://andreamaduzzi.github.io/llana/}}
        % <-this % stops a space

\maketitle

\begin{abstract}
Recent advances in Multimodal Large Language Models (MLLMs) have shown remarkable capabilities in understanding both images and 3D data, yet these modalities face inherent limitations in comprehensively representing object geometry and appearance. Neural Radiance Fields (NeRFs) have emerged as a promising alternative, encoding both geometric and photorealistic properties within the weights of a simple Multi-Layer Perceptron (MLP). This work investigates the feasibility and effectiveness of ingesting NeRFs into an MLLM. We introduce \model{}, the first MLLM able to perform new tasks such as NeRF captioning and Q\&A, by directly processing the weights of a NeRF's MLP. 
Notably, \model{} is able to extract information about the represented objects without the need to render images or materialize 3D data structures. 
In addition, we build the first large-scale NeRF-language dataset, composed by more than 300K NeRFs trained on ShapeNet and Objaverse, with paired textual annotations that enable various NeRF-language tasks.
%with no human intervention the first dataset of NeRFs with paired textual annotations enabling various NeRF-language tasks, built uponShapeNet. 
Based on this dataset, we develop a benchmark to evaluate the NeRF understanding capability of our method. Results show that directly processing NeRF weights leads to better performance on NeRF-Language tasks compared to approaches that rely on either 2D or 3D representations derived from NeRFs.
\end{abstract}
\section{Introduction}
\label{sec:intro}
The field of Natural Language Processing has been profoundly transformed by Large Language Models (LLMs)~\cite{kenton2019bert, raffel2020exploring, GPT4, llama}, due to their text comprehension and generation capabilities. These results have fostered the development of Multimodal LLMs (MLLMs)~\cite{Palm-E, LLaMA-Adapter, llava, InstructBLIP, VideoLLM}, which can process various modalities such as images, videos and audio, to generate text describing and reasoning about the content of such modalities. 
Recently, MLLMs have also been extended to 3D data~\cite{pointllm, gpt4point, 3dllm}, primarily represented as colored point clouds, yielding remarkable results even in this scenario.

%Although images and colored point clouds are rich representations of an object, they both miss some aspects of it. Indeed, while images capture the interaction of the object's surface with light and by definition provide photorealistic renderings of it, they do not make explicitly available the 3D structure of an object, which can only be recovered using further processing of multiple views. Vice versa,  point clouds do provide the 3D structure of an object but, even if they store an RGB value for each point, they do not store its photorealistic appearance. 
Another approach for representing objects and scenes has emerged alongside traditional images and 3D data: Neural Radiance Fields (\nerf{}s)~\cite{nerf}. \nerf{}s are coordinate-based neural networks, typically Multi-Layer Perceptrons (MLPs), designed to capture both the geometry and the photorealistic appearance of an object. By learning a continuous radiance field across 3D space, NeRFs can be used to generate realistic images from any viewpoint or reconstruct the object's 3D surface by querying the trained model.
Using \nerf{}s to represent 3D data offers distinct advantages over conventional approaches like multi-view images or point clouds. The continuous nature of \nerf{}s allows generating unlimited photorealistic images at any desired resolution while only requiring the storage of MLP weights rather than a large collection of images.
Due to their benefits, \nerf{}s are effectively becoming a new modality stored and communicated independently, with datasets of \nerf{}s being made publicly available~\cite{hu2023nerf, ramirez2023deep} and companies providing digital twins of objects represented as \nerf{}s. % (e.g., {\footnotesize \url{https://lumalabs.ai/}}).
 
The increasing adoption of NeRFs and their appealing characteristics prompted us to investigate the following research question: is it possible to build an MLLM able to directly ingest \nerf{}s? 
Inspired by recent studies on meta-networks that can process neural fields~\cite{ramirez2023deep, lim2024graph}, we answer this question positively by showing that it is possible to process the weights of a given \nerf{} with a meta-network encoder that projects the NeRF weights into the embedding space of a pre-trained LLM such as \llama{} 2~\cite{llama}. By doing so, we create the first MLLM for NeRFs, dubbed Large Language and \nerf{} Assistant (\model{}), which can perform \nerf-language tasks such as \nerf{} captioning, \nerf{} Q\&A and zero-shot \nerf{} classification. 

In the former version of this paper~\cite{amaduzzi2024llana}, we introduced ShapeNeRF-Text, the first \nerf{}--language dataset, comprising language annotations for 40K objects from ShapeNet.
To collect this dataset, we designed an automated annotation framework that leverages MLLMs to produce text annotations for \nerf{}s trained on 3D models. Using this dataset alongside an additional split containing manually curated textual descriptions \cite{amaduzzi2023looking}, we established a benchmark for \nerf{} textual assistants. 
Building upon such a foundation, this work introduces several key advances: first, we significantly expand the scale and diversity of NeRF-language understanding by introducing ObjaNeRF-Text, a new dataset of NeRFs, built upon Objaverse~\cite{deitke2023objaverse}. With 280K annotated NeRFs, this new dataset represents a seven-fold increase in scale compared to ShapeNeRF-Text. While ShapeNeRF-Text was limited to synthetic objects from 10 classes of ShapeNet with machine-generated annotations, ObjaNeRF-Text provides two key improvements: it enlarges the variety of synthetic objects and introduces real-world objects, while also incorporating high-quality human-written annotations from \cite{pointllm} and \cite{gpt4point}, providing a higher quality and more diverse benchmark for NeRF-language understanding.
Secondly, we extend our previous experimental setup by investigating the LLM scaling effects on NeRF-language tasks. These experiments provide valuable insights into how the size of the underlying LLM influences the performance of MLLMs in processing and understanding 3D neural fields.
When evaluating \model{}, we compare it against traditional approaches that process \nerf{}s by first converting them to explicit data representations -- either rendered images or 3D point clouds -- and then using existing MLLMs designed for these modalities. Through a comprehensive evaluation on our proposed benchmark, we demonstrate the advantages of our direct NeRF processing approach.
We show that the quality of MLLM outputs is adversely affected by both the resolution of extracted 3D geometry and images, as well as the choice of the viewpoint used for image rendering.
Important details might be lost by rendering from the wrong angle, or the extracted geometry might not be detailed enough. Vice versa, by operating directly on the MLP weights, we are able to extract all the information about the object without any other design decision. Our approach turns out to be the most effective way to create a NeRF assistant, as it consistently outperforms MLLMs processing images or 3D geometries extracted by querying \nerf{}s.

\begin{figure*}[t]
    \centering
    \includegraphics[width=0.95\linewidth]{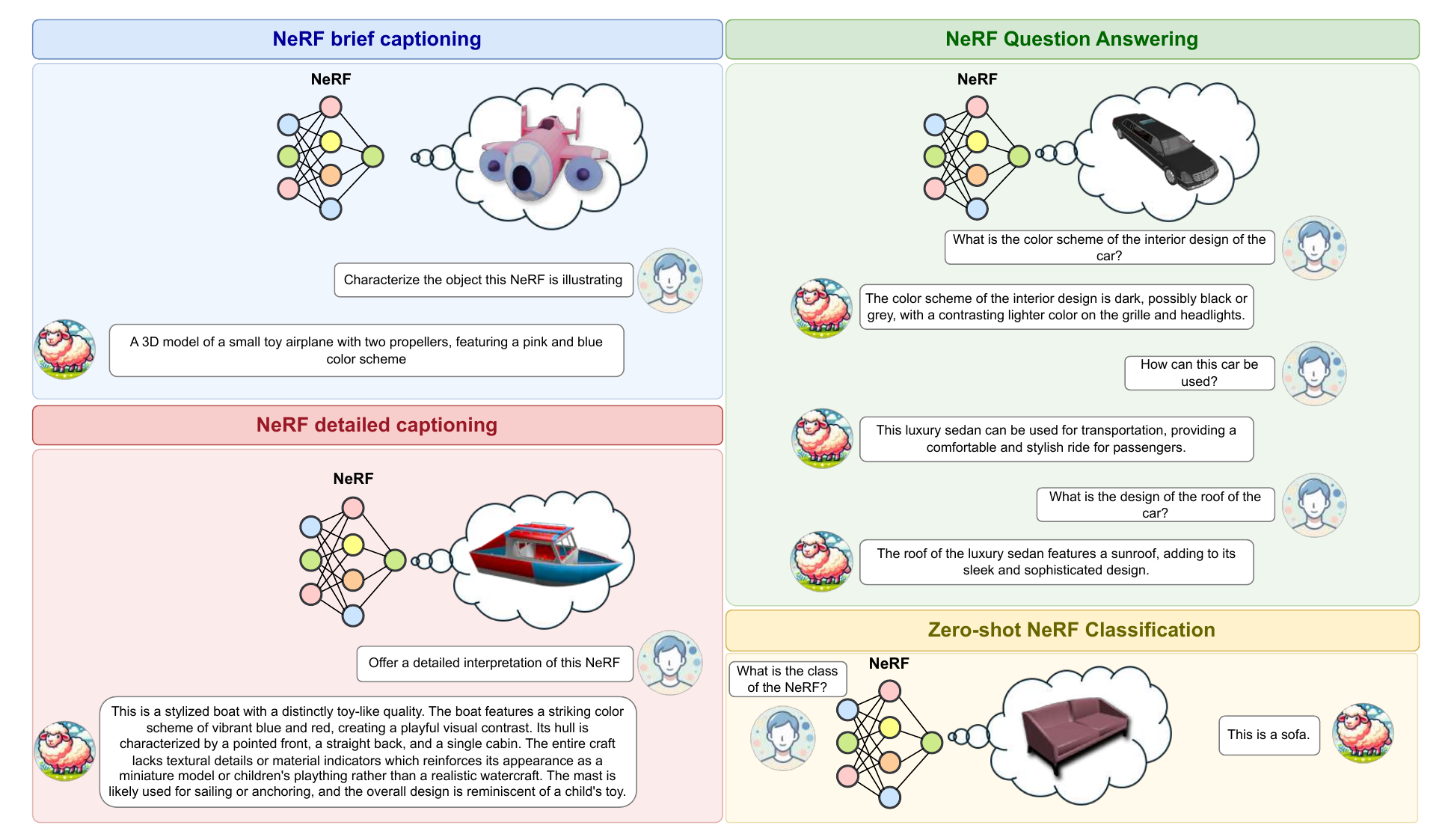}
    \caption{\textbf{\model{}.} A new Multimodal Large Language Model that understands and reasons on an input \nerf{}. Notably, our framework processes directly the \nerf{} weights and performs tasks such as captioning, Q\&A, and zero-shot classification of \nerf{}s.}
    \label{fig:teaser}
\end{figure*}

The key differences with~\cite{amaduzzi2024llana} are:
\begin{itemize}
\item{We create ObjaNeRF-Text, the largest existing NeRF-Language dataset, providing 280K NeRFs paired with textual annotations sourced from \cite{pointllm} and \cite{gpt4point}, with a seven-fold increase in scale over ShapeNeRF-Text \cite{amaduzzi2024llana}. This new dataset expands the variety of synthetic objects and incorporates real-world objects. Moreover, unlike the machine-generated textual annotations of ShapeNeRF-Text, the test set of ObjaNeRF-Text features high-quality human-written conversations, providing more natural and reliable ground-truth data.}
\item{We explore the impact of the LLM size on \nerf{} understanding by extending our previously proposed model \model{} to utilize LLAMA-13b, offering new insights into how LLM size affects performance on NeRF-language tasks.}
\end{itemize}

The summary of our contributions is:
\begin{itemize}
    \item{\model{}, the first MLLM capable of performing tasks such as captioning and Q\&A on \nerf{}s.}
\item{We show that it is possible to build such an assistant by directly processing the \nerf{}s weights with a meta-encoder, which is faster and captures more information compared to rendering images or extracting 3D data.}
\item{A NeRF-Language benchmark for MLLMs, built on ShapeNet and Objaverse, which contains more than 320K NeRFs of synthetic and real objects, paired with text annotations. Our evaluation on this benchmark demonstrates that \model{} outperforms traditional MLLMs operating on discrete representations derived from NeRFs.}
\item{An analysis of the impact of the LLM size on MLLMs evaluated on NeRF-Language tasks.}
\end{itemize}

%
%Thus, our proposed MLLM approach, namely Large Language and \nerf{} Assistant (\model{}), ingest the weights of a given \nerf{} with a meta-network encoder and then projects the network representation to the token input space of a pre-trained LLM such as Llama \cite{llama}, which yields a textual representation used for \nerf-language tasks such as \nerf{} captioning, Q\&A or Zero-shot \nerf{} classification (see \cref{fig:teaser}).
%

\section{Related work}
{\bf{Multimodal Large Language Models.}}
Significant advancements have been made by Large Language Models (LLMs) in language understanding, reasoning, and generalization capabilities~\cite{kenton2019bert, raffel2020exploring, GPT4, llama}.
These models have been extended into Multimodal Large Language Models (MLLMs), which broaden their reasoning abilities by including other modalities like images~\cite{Palm-E, LLaMA-Adapter, LLaMA-Adapterv2, image-bind}, audio~\cite{Audiogpt}, and videos~\cite{VideoChatGPT, VideoLLM}. MLLMs generally align target features with the corresponding textual ones and then incorporate them into LLMs to perform various text-based inference tasks. 
Some MLLMs are trained entirely from scratch~\cite{Kosmos-1, Kosmos-2}, others utilize pretrained LLMs~\cite{Otter, Qwen-VL, llava, BLIP-2, InstructBLIP}. 
3D MLLMs focus on understanding the 3D world typically represented in one of two ways: as colored point clouds~\cite{gpt4point, 3dllm, 3d-vista, Point-bind, pointllm} or multi-view images~\cite{Hong_2023_CVPR}. 
These models use different training approaches - some learn from 2D images~\cite{3dllm, 3d-vista, Hong_2023_CVPR}, while others are trained by directly matching text descriptions with points~\cite{Point-bind, pointllm, gpt4point}.

{\bf{Neural radiance fields.}}
\nerf{}~\cite{nerf} have been applied in several visual tasks such as novel view synthesis~\cite{martin2021nerf}, generative media~\cite{poole2022dreamfusion}, and robotics~\cite{yen2022nerf}.
%, and computational photography~\cite{mildenhall2022nerf}.
The base formulation employs MLPs to convert spatial coordinates into colors and densities. 
Recent advancements substitute or enhance MLPs with explicit data structures~\cite{Chen2022ECCV, sun2022direct, Plenoxels, instant} for faster training and inference. 
%We employ the same base \nerf{} formulation as ~\cite{nf2vec}, i.e., each \nerf{} is a single MLP extracting density and color information for each 3D coordinate. Each MLP represents a different object.   % \nftovec{}

{\bf{Neural radiance fields and language.}}
The interaction between \nerf{} and language has been recently investigated for several practical applications. Many works address the problem of generating geometrically consistent views of objects or scenes described by textual prompts~\cite{seo2023dittonerf, Metzer_2023_CVPR, Jo_2023_WACV, NEURIPS2023_a0303731, li2024instructpixnerf, lee2022understanding, poole2022dreamfusion}. 
Other approaches focus on editing the scene represented by a~\nerf{} through text, e.g., by changing the appearance and shape of objects~\cite{wang2022clip, Hwang_2023_ICCV, Song_2023_ICCV, 10144678, Sun_2024_WACV, Haque_2023_ICCV, 10476703, zhuang2023dreameditor}, or by inserting/removing objects in the scene~\cite{bai2023componerf, Mirzaei_2023_ICCV}.
Some techniques investigate new types of radiance fields that predict language features for each spatial location alongside density and color~\cite{lerf2023, NEURIPS2022_93f25021}. By transferring knowledge from vision-language models into these enhanced radiance fields, they can be queried by textual prompts.
Such \emph{language fields} are parametrized by a neural network.
%
%\textcolor{red}{\textbf{Unlike all previous methods, \citet{ballerini2024clip2nerf} is the first to utilize \nerf{}s as an input modality.}} 
Unlike all previous methods, the solution proposed in~\cite{ballerini2024clip2nerf} is the first to utilize the weights of a \nerf{}'s MLP as an input modality. 
This method aims to learn a mapping between the embedding spaces of the \nerf{} and CLIP~\cite{clip} to perform tasks such as \nerf{} retrieval from textual or image queries. 
%Similarly, we consider \nerf{}s as an input modality to our framework. 
Differently, our goal is to develop an MLLM capable of reasoning about  \nerf{}s.

{\bf{Deep learning on neural networks.}}
Several studies have explored using meta-networks, i.e., neural networks that process other neural networks. Initially, researchers concentrated on predicting network characteristics, such as accuracy and hyperparameters, by processing their weights~\cite{Unterthiner2020PredictingNN,urholt2021selfsupervised, knyazev2021parameter,jaeckle2021generating,Lu2020Neural}.
Several recent works focus on processing networks that implicitly encode data, e.g., Implicit Neural Representations (INR) or Neural Fields. These methods are able to classify or segment data by processing solely the weights of the input neural networks.
Functa~\cite{functa} trains a single shared network on a full dataset to learn compact modulation embeddings for each sample, which can then be used for various downstream tasks.
More recent research has shifted focus to analyzing networks that represent individual data samples, such as networks trained to model specific objects.
By leveraging a novel encoder architecture for MLP weights, \inrtovec{}~\cite{deluigi2023inr2vec} extracts compact embeddings from INRs of 3D shapes, which are employed as inputs for downstream tasks. 
\nftovec{}~\cite{ramirez2023deep} extends \inrtovec{} to ingest the \nerf's network weights to classify, segment, or retrieve similar \nerf{}s. 
%Notably, \nftovec{} can process even large MLPs thanks to the efficient design of the encoder. 
The solution from~\cite{cardace2024neural} develop a strategy to process neural fields represented by a hybrid tri-plane structure.  
Other approaches~\cite{navon2023equivariant, zhou2023neural, zhou2023permutation, zhou2024universal} develop equivariant architectures to handle MLPs by exploiting weight space symmetries~\cite{hecht1990algebraic} as an inductive bias. 
Also, Graph Neural Networks have been investigated to compute a network representation~\cite{kofinas2024graph, lim2024graph}. 
Since we aim to process \nerf s directly from the network weights, we employ \nftovec{} as our meta-encoder due to its efficient and scalable architecture.
\section{Methodology}
\label{sec:method}
This section describes the proposed Large Language and \nerf{} Assistant (\model{}). We first provide an overview of \nerf{}s and the meta-encoder that maps \nerf{} weights into a global embedding. Then, we present the overall \model{} framework and discuss our training protocol.

\begin{figure*}[!t]
    \centering
    \includegraphics[width=0.9\linewidth]{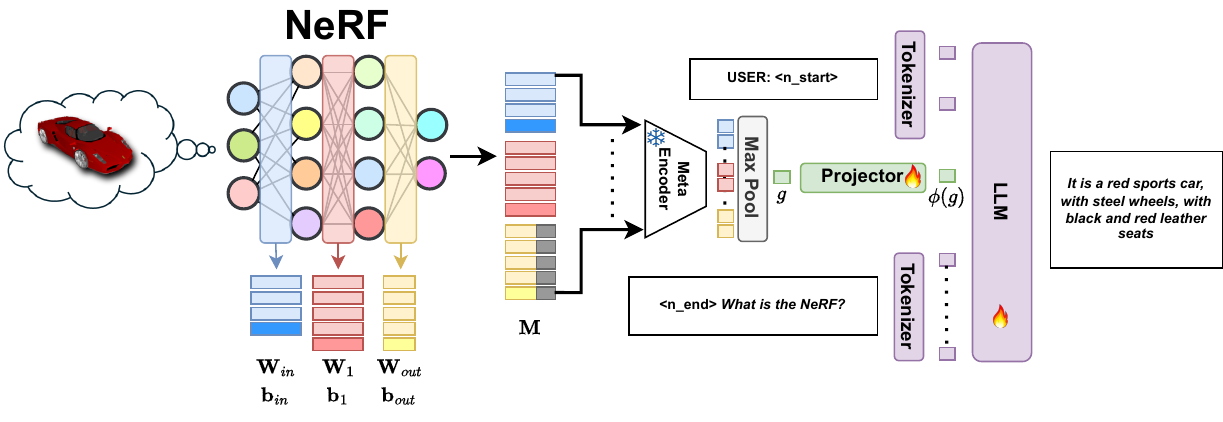}
    \captionsetup{justification=centering}
    \caption{\textbf{Framework overview. Example of \nerf{} captioning.}}
    \label{fig:framework}
\end{figure*}

{\bf{Neural Radiance Fields}}
%\label{sec:nerf}
A Neural Radiance Field (NeRF)~\cite{nerf} is a framework that employs coordinate-based neural networks, typically Multi-Layer Perceptrons (MLP), to model 3D scenes or objects. It is trained on a set of images taken from various vantage points. Once trained, the \nerf{} can be exploited to perform novel views synthesis, i.e., photorealistic rendering of images from viewpoints unseen at training time.

In its base formulation, the MLP is a function of continuous 3D coordinates $\vb{p} = (x, y, z) \in \mathbb{R}^3$, that yields four-dimensional outputs, $RGB\sigma \in [0,1]^4$. 
This output encodes the $RGB$ color and the volume density $\sigma$ at each 3D location in the scene. The volume density $\sigma$ can be interpreted as the differential probability of a ray terminating at a point $\vb{p}$. 
After training, a \nerf{} can render images from any desired viewpoints at arbitrary resolution by querying it for the values of $RGB$ and $\sigma$ at several points along the ray corresponding to each pixel and applying the volumetric rendering equation~\cite{nerf}.

In this work, we implement \nerf{}s as MLPs composed of $L$ hidden layers, an input layer, and an output layer. An example of MLP with 1 input, 1 output, and 1 hidden layer is shown in~\cref{fig:framework} (left).
A layer is parameterized by a weight matrix plus a bias vector. 
More in detail, the hidden layers in our architecture have the same number of input and output neurons, $H$, thus having squared weight matrices $\mathbf{W}_{l} \in \mathbb{R}^{H \times H}$ for $l=1,\dots,L$ and $H$-dimensional biases $\mathbf{b}_l \in {R}^{H}$. 
The input $\vb{p}$ goes through a $24$-frequency encoding~\cite{nerf}, therefore the first layer has $\mathbf{W}_{in} \in \mathbb{R}^{144 \times H}$ and $\mathbf{b}_{in} \in \mathbb{R}^{H}$. 
The final layer has $\mathbf{W}_{out} \in \mathbb{R}^{H \times 4}$ and $\mathbf{b}_{out} \in \mathbb{R}^{4}$. 
Adopting the same architecture used in~\cite{ramirez2023deep}, an instance of the employed \nerf{} has $L=3$ hidden layers, with $64$ neurons each. The $ReLU$ activation function is applied between all layers except for the last one, which directly computes the density and $RGB$ values without any activation function. 
%A frequency encoding~\cite{nerf} is applied to the input 3D coordinates, in order to improve the \nerf{} reconstruction quality. 
Our \nerf{}s are trained using a Smooth $L_1$ loss \cite{girshick2015fast} between the predicted and ground-truth $RGB$ pixel intensities, weighting background pixels less than foreground pixels ($0.8$ foreground vs. $0.2$ background). 
%In this process, the final rendered image is obtained by accumulating colors and opacities through volumetric rendering~\cite{nerf}.
The final rendered images are obtained by querying the neural network with 3D coordinates to obtain RGB color values and density estimates. 
These values are then integrated along camera rays using volumetric rendering techniques~\cite{nerf}, to produce the final image. Each \nerf{} is trained for approximately $2000$ steps, until it achieves good reconstruction quality as measured by the Peak Signal-to-Noise Ratio (PSNR).
%
%Due to their many advantages over other data representations, NeRFs have become a new standard for describing the world. They have become a new input modality, stored and communicated independently.

{\bf{Meta-encoder}}
%\label{sec:meta-encoder}
In this work, we investigate how to design a Multimodal Large Language Model (MLLM) that works directly on the weights of \nerf{}s. We expect the \nerf{} weights to contain comprehensive information about the represented object, such as its geometry and appearance. Thus, an encoder could extract all the relevant information from these weights to perform language-based tasks like generating captions and answering questions about the object.

Inspired by the recent development of meta-networks capable of processing neural fields~\cite{lim2024graph, ramirez2023deep}, we employ \nftovec{} \cite{ramirez2023deep} as our meta-encoder architecture. 
This approach takes as input the weights of a \nerf{} and provides as output a global embedding that distills the content of the input. 
In particular, the weight matrices and biases of the input \nerf{} are stacked along the row dimension to form a matrix $\mathbf{M} \in \mathbb{R}^{S \times H}$, where $S= 144 + 1 + L*(H+1) + H + 1 = L*H + L + H + 146$. Before stacking, we pad the weights and biases of the output layer, $\mathbf{W}_{out}$ and $\mathbf{b}_{out}$, with zeros to obtain $H$ columns (see \cref{fig:framework}, center).

The meta-encoder is parametrized as an MLP with batch normalization layers~\cite{ioffe2015batch} and $ReLU$ non-linearities. To gracefully scale with the MLP input dimensions, the encoder processes each row of $\mathbf{M}$ independently, extracting a total of $S$ tokens, each of length $G$, from an input \nerf{}. 
Then, they are processed by a max-pooling layer to provide a global representation $g \in  \mathbb{R}^{G}$ of the \nerf{}, with $G=1024$ in our experiments.
The encoder has been pre-trained on the \nerf{}s from ShapeNeRF--Text and ObjaNeRF--Text by applying the self-training protocol of \nftovec{}~\cite{ramirez2023deep}, i.e., jointly with a decoder architecture that, given as input the \nerf{} global embedding, reconstructs the same images as the input \nerf{} from arbitrary viewpoints.

{\bf{Large Language and \nerf{} Assistant}}
%\label{sec:llana}
Inspired by recent approaches that proposed effective Multimodal Large Language Models, we build \model{} by leveraging on a pre-trained LLM with a Transformer backbone~\cite{transformers}, in our experiments \llama{}~2~\cite{llama}, and projecting the NeRF modality into its embedding input space, as proposed for images and 3D data~\cite{llava, pointllm} (see \cref{fig:framework}, right). 
Thanks to the self-attention mechanism, the transformer can understand the contextual relationships between the text and the \nerf{} tokens, enabling it to generate responses based on both the text and the \nerf{} inputs. 

We define a projector network, $\phi$, composed of a stack of $3$ trainable linear layers, interleaved with $GeLU$ activation functions, that project the embedding of the input \nerf{} computed by the meta-encoder into the embedding space of \llama{} 2. 
More in detail, the \nerf{} embedding is encapsulated between two special tokens, \texttt{<n\_start>} and \texttt{<n\_end>}, whose embeddings are learned end-to-end while training. 

%THen, given an input sequence of mixed \nerf{} and word tokens, $(\texttt{<n\_start>},\phi(g), \texttt{<n\_end>}, w_1, w_2, ..., w_k)$, where $k$ is the number of word tokens, the LLM returns a sequence of predicted word tokens $(\hat{w}_{k+1}, \hat{w}_{k+2}, \dots, \hat{w}_{eos})$.
Finally, an input sequence composed by the \nerf{} embedding and $k$ word tokens, $(\texttt{<n\_start>},\phi(g), \texttt{<n\_end>}, w_1, w_2, ..., w_k)$, is provided as input to the LLM which predicts a sequence of word tokens $(\hat{w}_{k+1}, \hat{w}_{k+2}, \dots, \hat{w}_{eos})$.
 
%Each mapped token is transformed into a probability distribution over a word vocabulary, and the prediction is the word with the highest probability.

{\bf{Training protocol}}
%\label{sec:training}
To train our framework, we hold multiple conversations about each \nerf{} leveraging both our ShapeNeRF--Text and ObjaNeRF--Text datasets (see \cref{sec:dataset}). 
These conversations are organized into a set of prompts from the user and expected ground-truth answers that are used to optimize the original auto-regressive objective of the LLM. 
For the meta-encoder, we train \nftovec{} on the \nerf{}s of ShapeNeRF--Text and ObjaNeRF--Text for 100 epochs on 4 NVIDIA A100 GPUs.
%During the training of \nftovec{}, we apply data augmentation to the NeRFs of ShapeNeRF--Text, incorporating color variations and geometric deformations.  
%The \nftovec{} weights are kept frozen when training \model{}. 
%In particular, \model{} follows a two-stage training protocol:
When training \model{} we follow a two-stage training protocol, keeping the \nftovec{} weights frozen:

%$\bullet$ 
\emph{Stage1: projector training.}
In the first stage, we train the projector network $\phi$ to align the \nerf{} and the word embedding spaces while keeping the LLM weights fixed. We train on an instruction dataset of brief descriptions from ShapeNeRF--Text and ObjaNeRF--Text to learn the projection layer efficiently. We also train the embeddings of the special tokens used to encapsulate the \nerf{} one. We optimize the projector weights and the embeddings for $3$ epochs with a learning rate of $0.002$ and a batch size of $16$ on each GPU.
        
%$\bullet$ 
\emph{Stage2: instruction tuning.} The second stage of training focuses on teaching the model to understand and reason about NeRF data using three types of text from ShapeNeRF--Text and ObjaNeRF--Text: brief descriptions, detailed descriptions, and Q\&A conversations. In this phase, we optimize both the projector and the LLM for $3$ epochs. We employ a learning rate of $0.00002$ and a batch size of $4$ on each GPU.

Our model is implemented in PyTorch and trained on NVIDIA A100 GPUs with 64GB of VRAM each. The model variant based on the 7B LLAMA architecture requires $4$ GPUs for training, while our largest version, which uses the 13B LLAMA architecture, needs $8$ GPUs. Training either version of the model takes approximately one day to complete.
%As for the safeguards of the LLMs employed, both LLaVA2-7b and Llama3-8b-chat are equipped with such mechanisms. 
%When fine-tuning \llama{} 2 for our specific applications, we strictly use a dataset that does not come from scraped web data. This approach should ensure that the integrity and effectiveness of the pre-existing safeguards are maintained.

%We train \model{} on a NeRF-Language dataset, ShapeNeRF-Text, described in \cref{sec:dataset}. In Stage 1, we train the model, . Then, in Stage 2, we optimize LLaMA 2 and the projector for 3 epochs on detailed descriptions, single-round and multi-round Q\&A with a learning rate of $0.0002$ and batch size 16. 
%\clearpage
\begin{figure*}[t]
    \centering
    \caption{\textbf{ObjaNeRF--Text statistics of ground-truth text annotations}}
    \setlength{\tabcolsep}{1pt}
    \begin{tabular}{c}
    \begin{tabular}{cccc}
    \multicolumn{4}{c}{\small Brief and Detailed Descriptions - Word clouds} \\
    \includegraphics[width=0.25\linewidth]{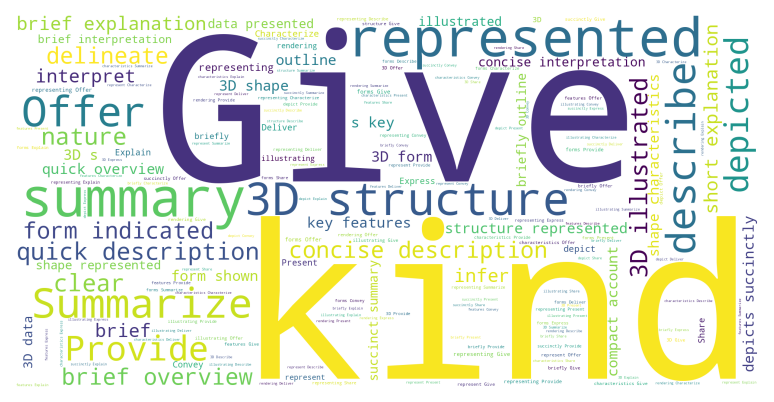} & \includegraphics[width=0.25\linewidth]{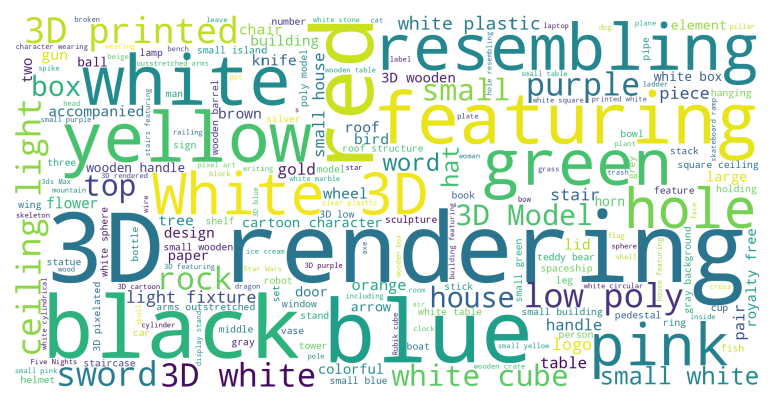}  &
    \includegraphics[width=0.25\linewidth]{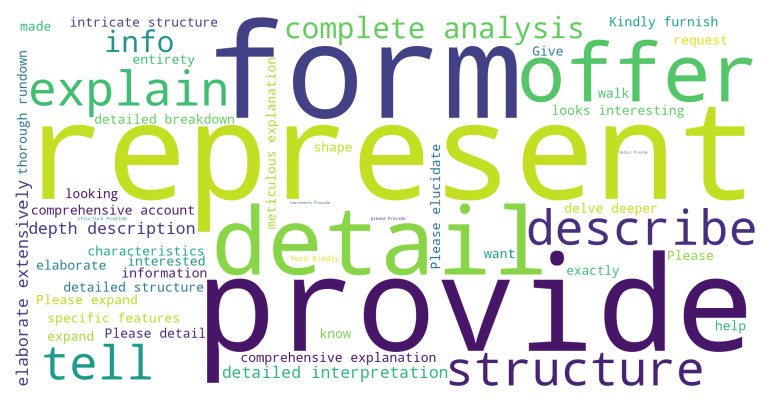} & \includegraphics[width=0.25\linewidth]{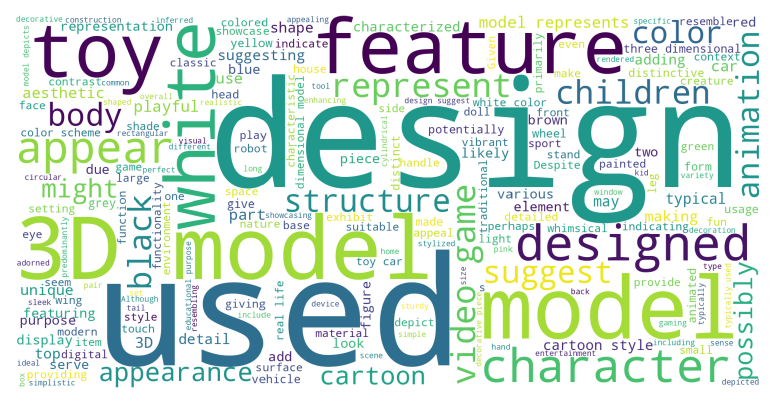} \\
    {\small Instructions (Brief)} & {\small Responses (Brief)} & {\small Instructions (Detailed)} & {\small Responses (Detailed)} \\
    & & & \\
    \multicolumn{4}{c}{\small Brief and Detailed Descriptions - Lengths (Words)} \\
    \includegraphics[width=0.25\linewidth]{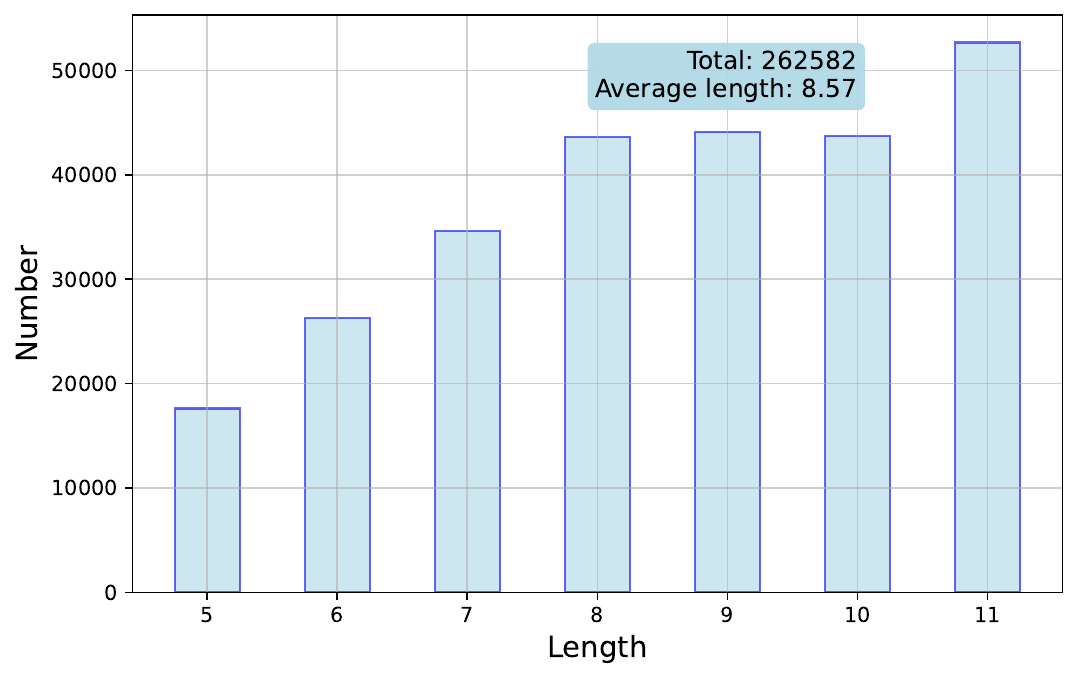} & \includegraphics[width=0.25\linewidth]{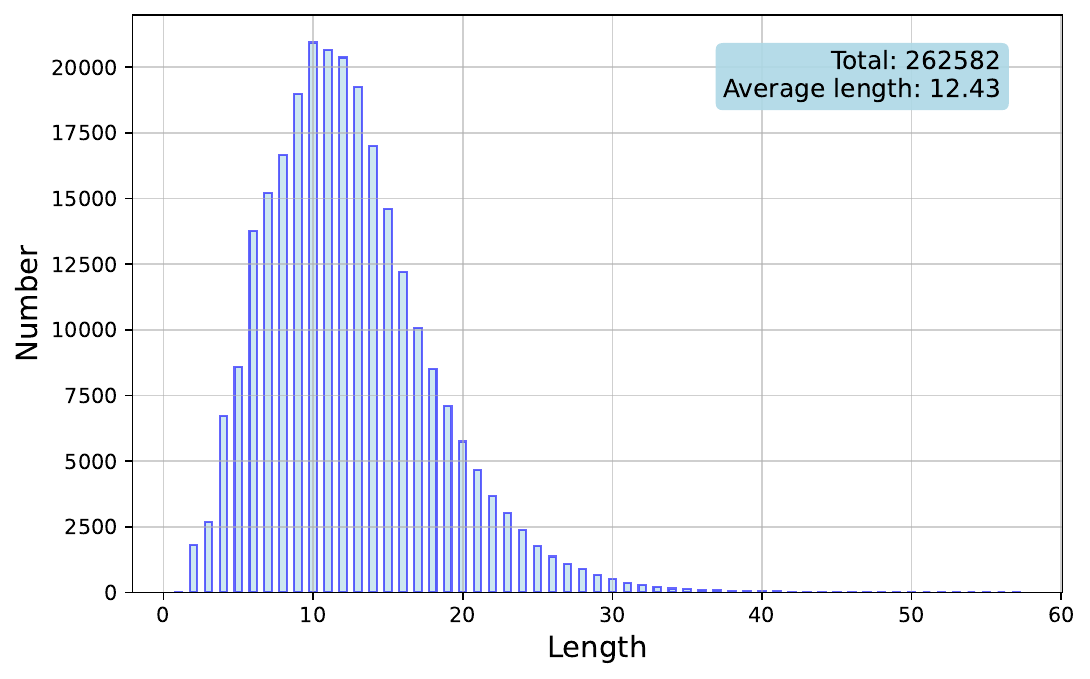} &
    \includegraphics[width=0.25\linewidth]{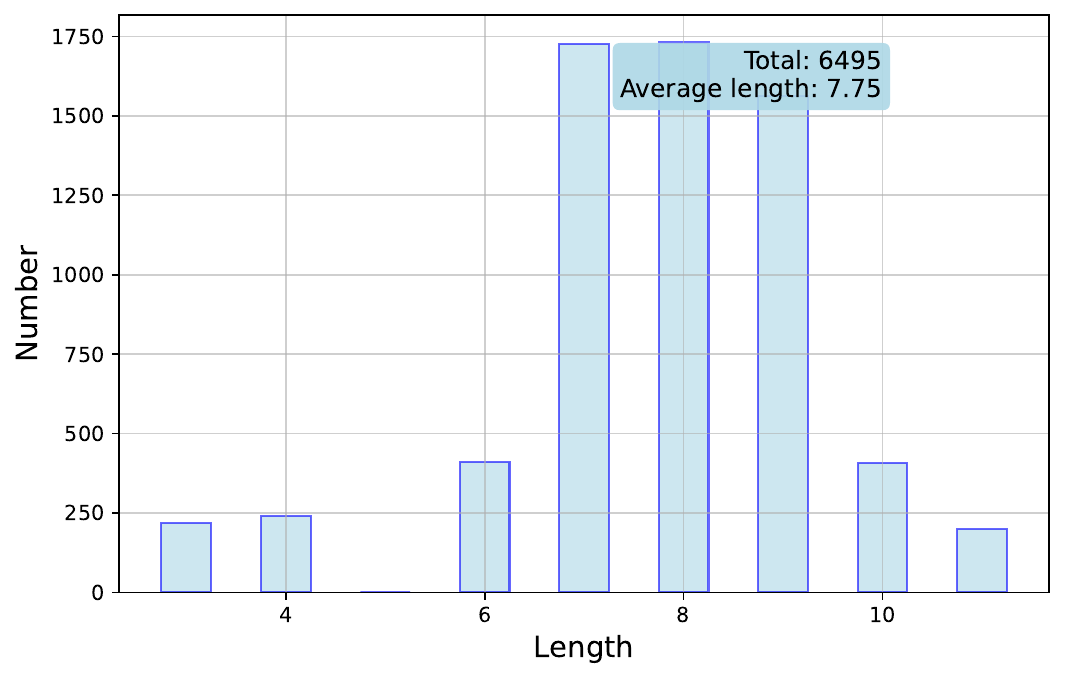} & \includegraphics[width=0.25\linewidth]{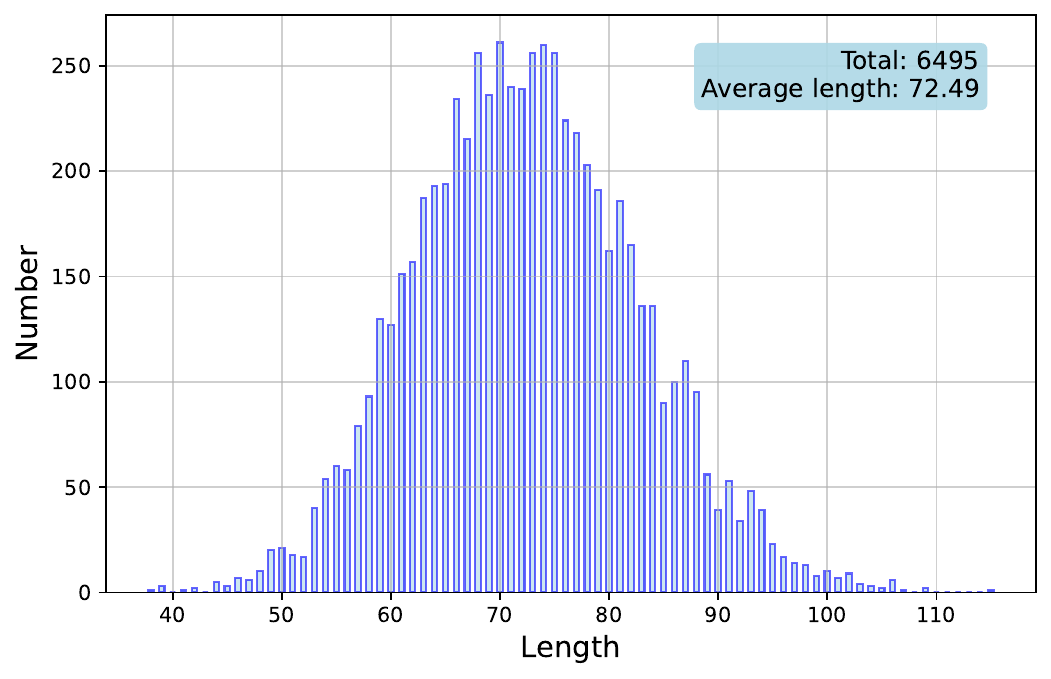} \\
    {\small Instructions (Brief)} & {\small Responses (Brief)} & {\small Instructions (Detailed)} & {\small Responses (Detailed)} \\
    & & & \\
    \multicolumn{4}{c}{\small Single-round and Multi-round Q\&A - Word clouds} \\
    \includegraphics[width=0.25\linewidth]{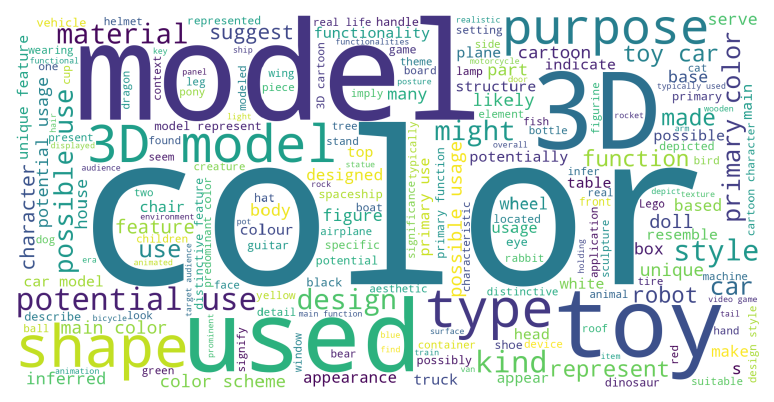} & \includegraphics[width=0.25\linewidth]{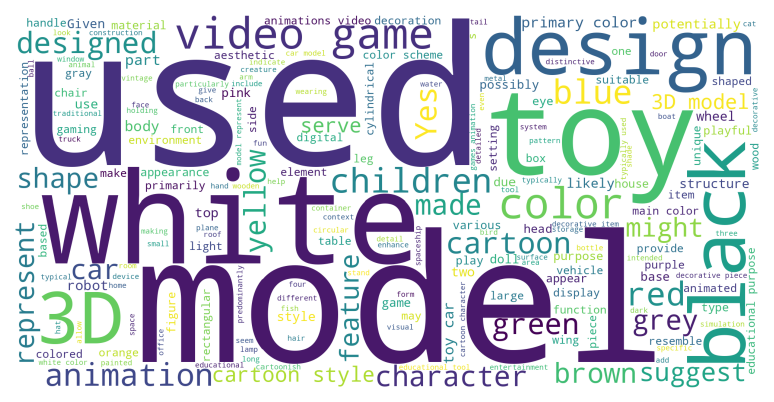} &
    \includegraphics[width=0.25\linewidth]{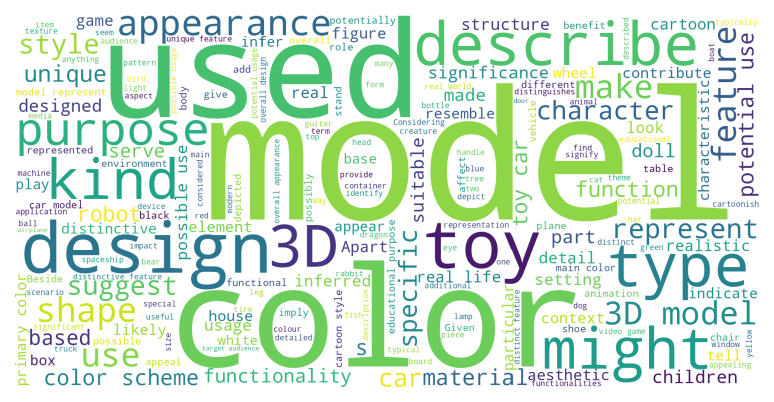} & \includegraphics[width=0.25\linewidth]{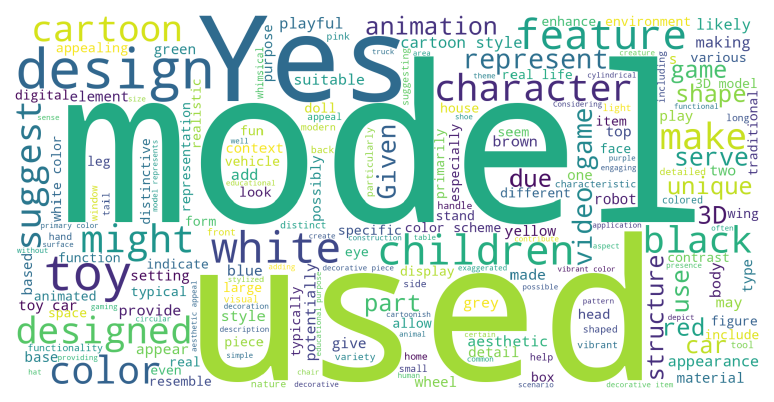} \\
    {\small Instructions (single-round)} & {\small Responses (single-round)} & {\small Instructions (multi-round)} & {\small Responses (multi-round)}\\
    & & & \\
    \multicolumn{4}{c}{\small Single-round and Multi-round Q\&A - Lengths (Words)} \\
    \includegraphics[width=0.25\linewidth]{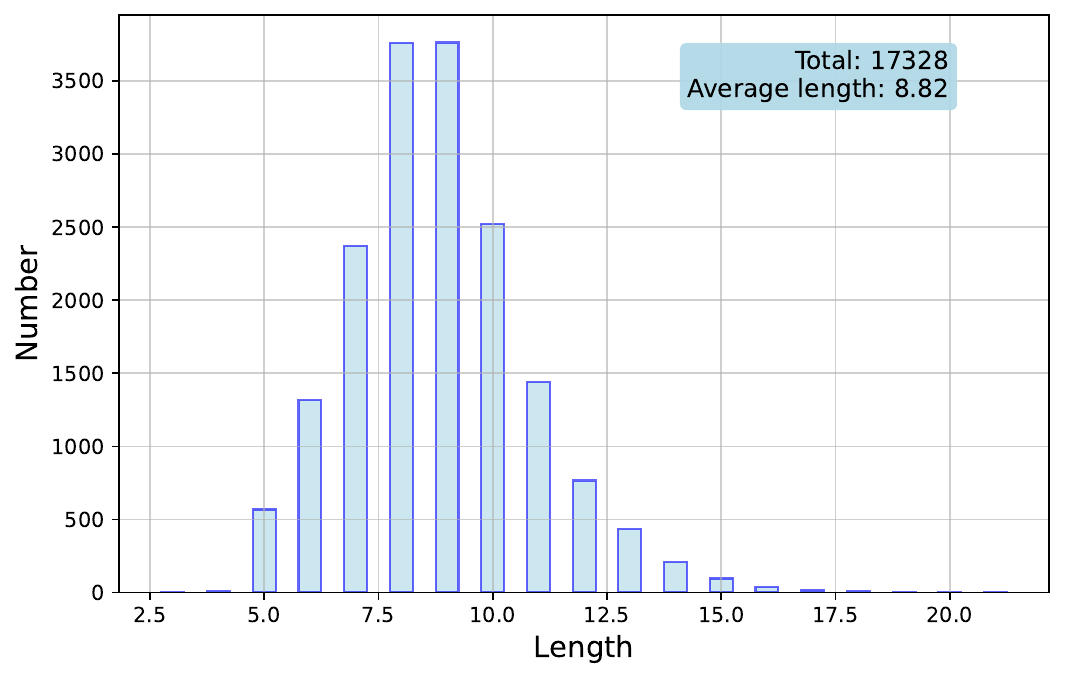} & \includegraphics[width=0.25\linewidth]{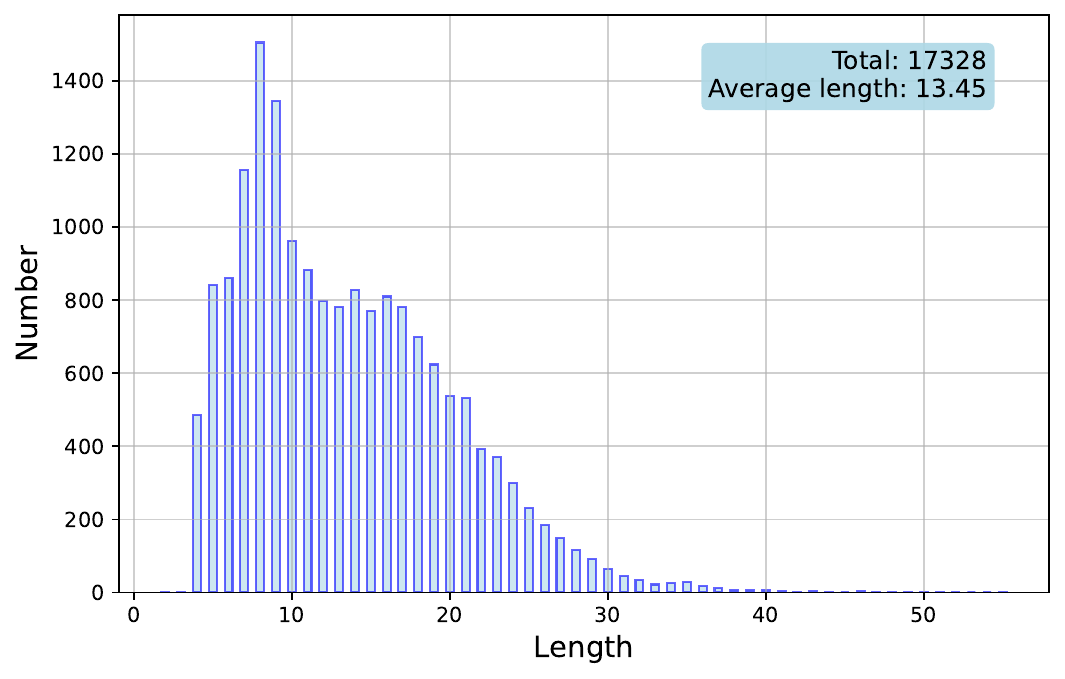} &
    \includegraphics[width=0.25\linewidth]{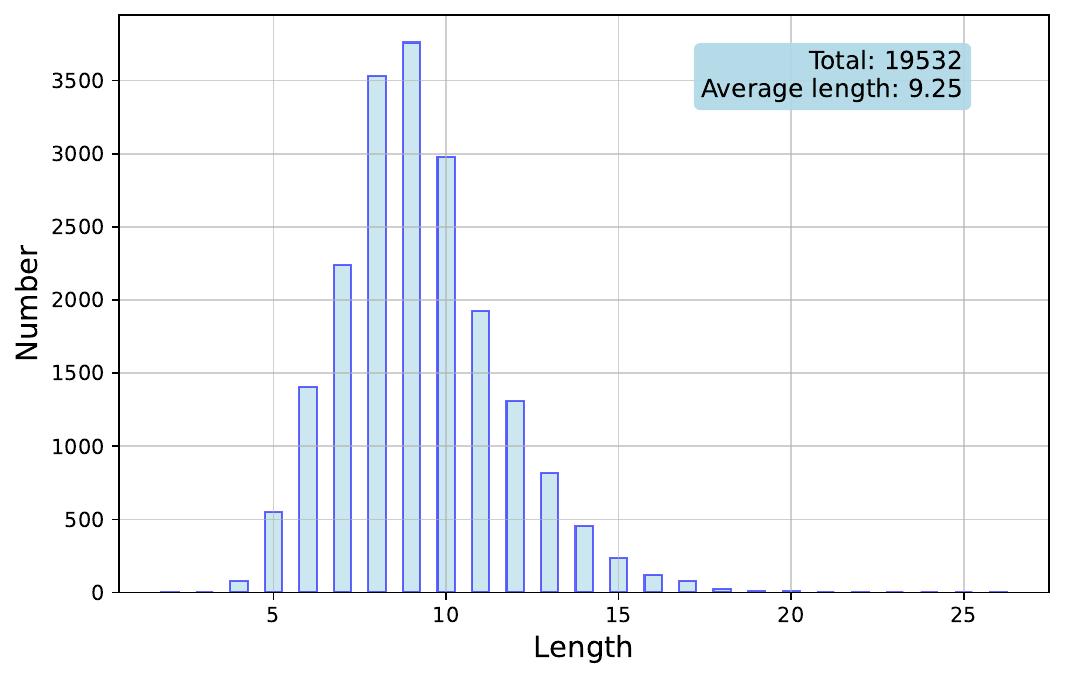} & \includegraphics[width=0.25\linewidth]{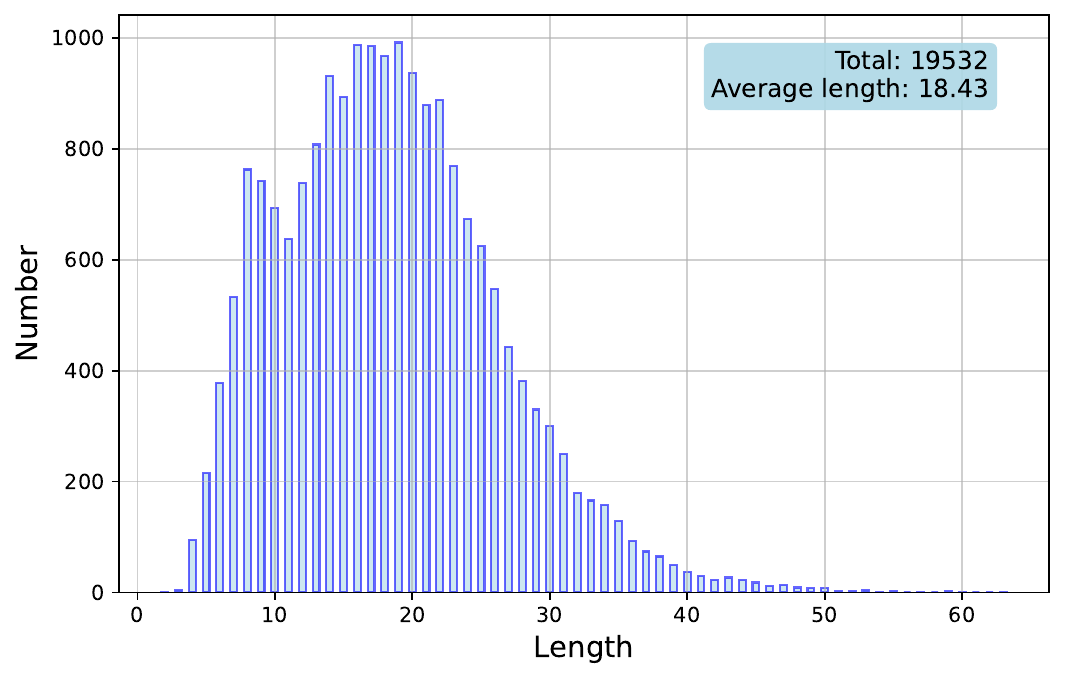} \\
    {\small Instructions (single-round)} & {\small Responses (single-round)} & {\small Instructions (multi-round)} & {\small Responses (multi-round)} \\
    \end{tabular} 
    \end{tabular}
    \label{tab:dataset_stats}
\end{figure*}

\section{Benchmark}
\label{sec:dataset}
To train and validate our NeRF assistant, we created a dataset of \nerf{}s with textual annotations. 
It features objects from ShapeNet~\cite{chang2015shapenet} and Objaverse~\cite{deitke2023objaverse}.

\subsection{ObjaNeRF--Text dataset}
\label{sec:objanerf}

%{\bf{\nerf{} data}} 
ObjaNeRF--Text is built leveraging on the rendered views from G-Buffer Objaverse~\cite{zuo2024sparse3d}. 
This dataset provides high-quality rendered views of a subset of 280K models from Objaverse. 
Each object was captured from $40$ different camera positions: $38$ views taken around the object at two different elevation angles, plus one view from above and one from below.
We train a \nerf{} on each object, leveraging these rendered views and following the procedure detailed in \cref{sec:method}

%The used \nerf{} architecture consists of a multi-layer perceptron (MLP) with $3$ hidden layers of $64$ neurons each. 
%ReLU activation functions are used between all layers except the output layer, which directly produces density and RGB values without any activation. 
%A frequency encoding~\cite{nerf} is applied to the input 3D coordinates, in order to improve the \nerf{} reconstruction quality. 
%Unlike the original \nerf{} architecture~\cite{nerf}, this version does not use the view direction as input. Instead, the MLP processes an input coordinate $\vb{p} \in \mathbb{R}^3$, to produce a 4-dimensional vector containing $RGB\sigma$.
%A \nerf{} is trained on each object, by minimizing the $L_1$ loss between the predicted and ground-truth RGB values, with different weights assigned to foreground ($0.8$) and background ($0.2$) pixels. 

\subsection{ShapeNeRF--Text dataset}
\begin{figure}[t]
    \centering
    \includegraphics[width=0.70\linewidth]{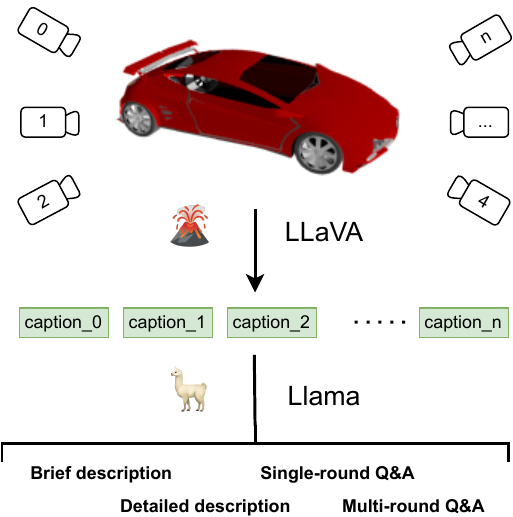}
    \caption{\textbf{Automatic annotation pipeline.} Given a 3D model, $N$ views are rendered and processed by a VLM (LLaVA) to generate view-specific captions. These are aggregated by an LLM (LLaMA) for final descriptions and Q\&A.}
    \label{fig:dataset}
\end{figure}

The textual annotations of ObjaNeRF--Text are derived from the dataset proposed in~\cite{pointllm}. This work provides machine-generated textual annotations for Objaverse models, divided into three categories: brief descriptions, detailed descriptions and Q\&A conversations. 
The brief descriptions are concise captions of the object, taking into account its global structure and appearance. 
Detailed descriptions are longer sentences that describe all the details of the object. 
The single-round Q\&As consist of a question about the object and the corresponding ground-truth answer, while the multi-round Q\&As are longer conversations formed by 3 questions and the relative answers. 
Objaverse provides colored meshes of both synthetic and real-world scanned objects.
When building ObjaNeRF--Text, we identified the intersection between the 280K objects in G-Buffer Objaverse and the 3D models from this textually annotated dataset. 
Overall, the training set of ObjaNeRF--Text comprises around 280K 3D models with brief descriptions, and 30K complex text annotations including detailed descriptions and Q\&A conversations, associated with 7K different objects.
For our evaluation benchmark, we created two different test sets: one for comparison with PointLLM~\cite{pointllm} and another for comparison with GPT4Point~\cite{gpt4point}, leveraging the splits proposed in these works. 
This dual\-split approach was necessary to ensure the largest and most fair evaluation possible since many samples in the test set of PointLLM were used to train GPT4Point. 
The final test sets contain human-annotated captions for $1366$ objects in the PointLLM test set and $518$ objects in the GPT4Point test set. 
To facilitate the training and test of \model{}, the words of the original text annotations referring to the point cloud data structure, such as ``point cloud'', ``3D point cloud'' or ``cloud of points'' have been modified into ``\nerf{}''.

{\bf{Statistical details of ObjaNeRF--Text}}
As for the training set, the average lengths in words for the instructions/responses are $8.57/12.43$ for brief descriptions, $7.75/72.49$ for detailed descriptions, $8.82/13.45$ for single-round Q\&As and $9.25/18.43$ for multi-round Q\&As.
\cref{tab:dataset_stats} shows histograms of the instruction/response length and the word clouds obtained after removing generic words such as ``model'', ``object'' and ``NeRF'', emphasizing frequent words in the ground-truth instructions and responses.

As quantitatively assessed in \cref{sec:language_only}, many of the questions belonging to the Q\&A set require a holistic 3D understanding of the object, to be answered correctly.
%We split the dataset into train, validation, and test sets, stratified on the ShapeNet classes, containing distinct objects in proportions of 80\%, 10\%, and 10\% of the total number of shapes, respectively.

ShapeNeRF--Text is the NeRF-Language dataset proposed in~\cite{amaduzzi2024llana}. This dataset provides 40K \nerf{}s of objects from ShapeNet, paired with language annotations. These ground-truth conversations have been automatically generated by leveraging LLaVA~\cite{llava} and LLAMA~\cite{llama}. 
More in detail, a \emph{brief description}, a \emph{detailed description}, 3 \emph{single-round Q\&As}, and one \emph{multi-round Q\&A} have been generated for every object. 
%The characteristics of these textual annotations are the same as those of ObjaNeRF--Text since
Our automatic data annotation pipeline is inspired by Cap3D~\cite{luo2024cap3d}.  
First, multiple views of each ShapeNet object have been rendered from different perspectives. 
Then, each view has been provided as input to \llava{} (\llava{}2-13b)~\cite{llava} to get a detailed description of the object from that point of view. 
Subsequently, starting from the captions generated by \llava{}, \llama{} 3 (\llama{}3-8B-chat) was used to generate the final ground-truth text data. 
An overview of this process is provided in \cref{fig:dataset}.

\subsection{Language tasks and metrics}
\label{sec:benchmark}
The experimental results on ShapeNeRF--Text and ObjaNeRF--Text are reported divided by tasks: brief captioning, detailed captioning and single-round Q\&A. Since the test sets of ObjaNeRF--Text contain short human-annotated captions, such results fall into the brief captioning category. Furthermore, for this task, we also evaluate the methods on the GPT2Shape Human Shape Text (HST) dataset~\cite{amaduzzi2023looking}, a subset of ShapeNet for which human-curated brief descriptions are publicly available. 
We employ standard language similarity metrics to evaluate these methods. 
We compute the cosine similarity between the global embeddings of the generated and ground-truth sentences provided by the pre-trained encoders Sentence-BERT~\cite{sentencebert} and SimCSE~\cite{simcse}. 
These metrics based on learned networks are the most effective in measuring the quality of the generated output~\cite{pointllm}. We also include standard handcrafted metrics based on n-gram statistics, like BLEU-1~\cite{bleu}, ROUGE-L~\cite{rouge}, and METEOR~\cite{meteor}.
\section{Experimental results}
As our method is the first to investigate language tasks on \nerf{}, there are no baselines in the literature. However, given a \nerf{}, a straightforward way to create an assistant for it could be to render an image and use a MLLM capable of ingesting images. Alternatively, we could extract the 3D shape from the \nerf{} and use one of the recent 3D MLLMs.  Hence, we evaluate \llava{} (v1.6)~\cite{llava} and BLIP-2~\cite{blip2} for images, as well as PointLLM~\cite{pointllm} and GPT4Point \cite{gpt4point} for colored point clouds. Since NeRFs can render arbitrary viewpoints after training, we also include the evaluation of LLaVA~\cite{llava} in a multi-view scenario. More in detail, we render images from $N$ viewpoints randomly sampled between the set of camera poses used to train each NeRF; then, we concatenate tokens from these N images and fed them into LLaVA alongside text instructions. We set $N$=$3$ because the model cannot process a higher number of images correctly. In addition, we test 3D-LLM~\cite{3dllm}, which processes meshes and multi-view images. When evaluating the baselines on ShapeNeRF-Text and ObjaNeRF-Text, we employ the official code and pre-trained models released by the respective authors\footnote{
\llava{}: \url{https://github.com/haotian-liu/LLaVA}
BLIP-2: \url{https://github.com/salesforce/LAVIS/tree/main/projects/blip2}
PointLLM: \url{https://github.com/OpenRobotLab/PointLLM}
GPT4Point: \url{https://github.com/Pointcept/GPT4Point}
3D-LLM: \url{https://github.com/UMass-Foundation-Model/3D-LLM}}. 
%We note that the only official GPT4Point weights available at submission time were those obtained from fine-tuning OPT-2.7B on Cap3D~\cite{cap3d}.
\subsection{Experiments on ObjaNeRF-Text and ShapeNeRF-Text}

Tables~\ref{tab:brief_frozen}, \ref{tab:brief_frozen_hst}, \ref{tab:detailed_frozen},  \ref{tab:qa_frozen} and \ref{tab:classification_frozen} show the results on ShapeNeRF--Text, while \cref{tab:objanerf_pointllm_test,tab:objanerf_gpt4point_test} show the results on ObjaNeRF--Text. 
The textual annotations of the ObjaNeRF--Text test sets consist of short descriptions, making them suitable for the evaluation of the brief captioning task.
In all tables, the baselines are ordered by increasing the size of the underlying LLM.

\begin{figure*}[t]
    \centering
    \includegraphics[width=\linewidth]{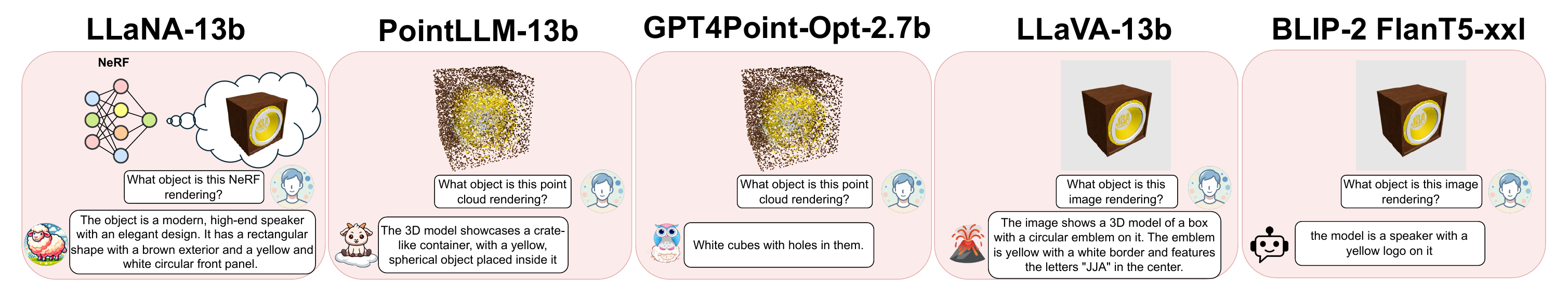}
    \caption{\textbf{Qualitative results on ShapeNeRF--Text brief descriptions.}}
    \label{fig:qualitative_results_shapenerf_brief}
\end{figure*}

\begin{figure*}[t]
    \centering
    \includegraphics[width=\linewidth]{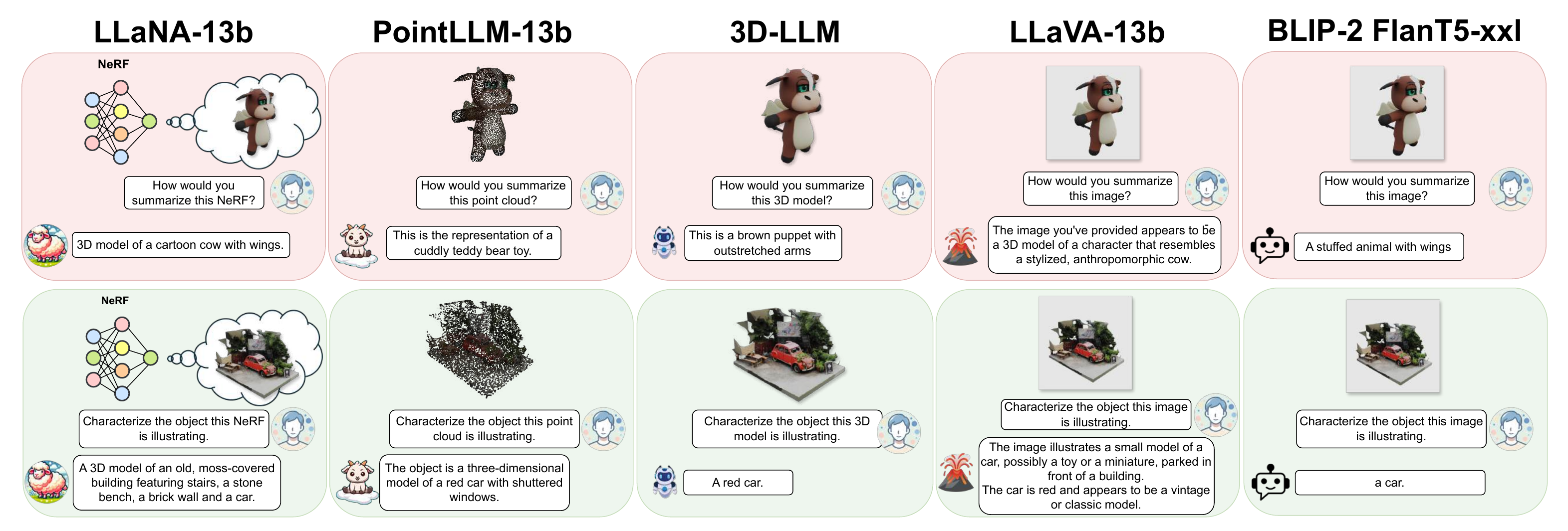}
    \caption{\textbf{Qualitative results on ObjaNeRF--Text brief descriptions  (PointLLM test set).}}
    \label{fig:qualitative_results_objanerf_pointllm}
\end{figure*}

\begin{figure*}[t]
    \centering
    \includegraphics[width=\linewidth]{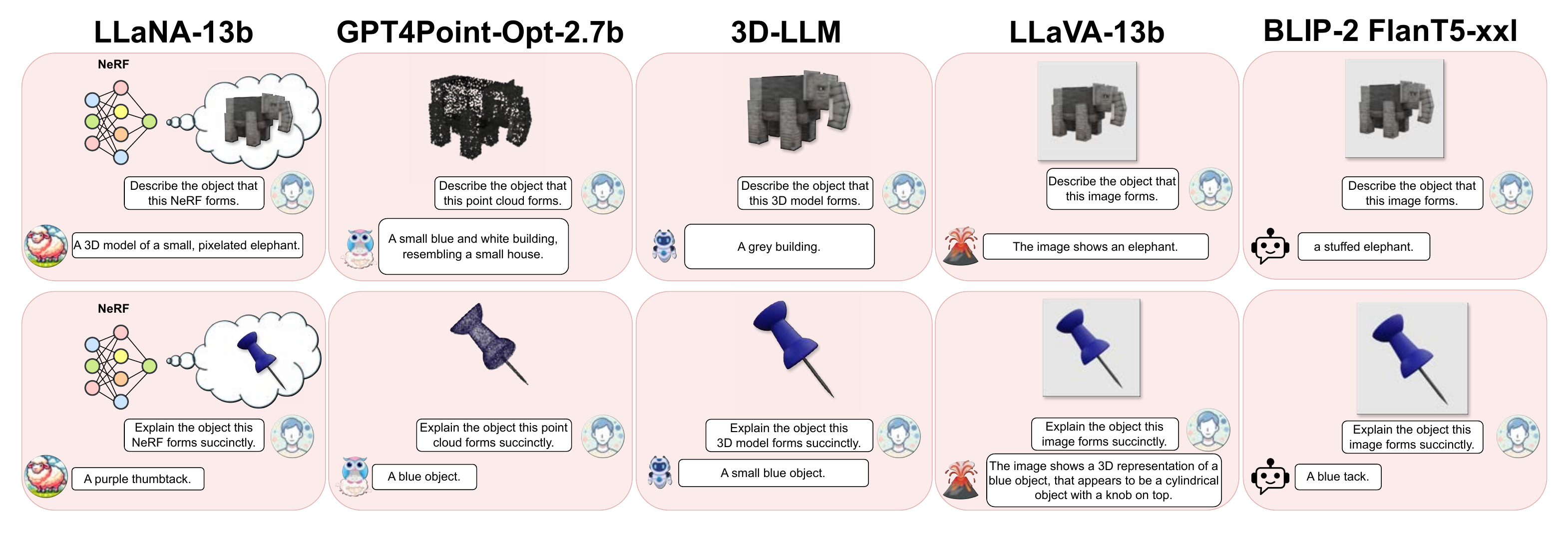}
    \caption{\textbf{Qualitative results on ObjaNeRF--Text brief descriptions (GPT4Point test set).}}
    \label{fig:qualitative_results_objanerf_gpt4point}
\end{figure*}

\paragraph{Brief captioning}
We report the results for the brief description task on ShapeNeRF--Text, HST and ObjaNeRF--Text in \cref{tab:brief_frozen}, \cref{tab:brief_frozen_hst}, \cref{tab:objanerf_pointllm_test} and \cref{tab:objanerf_gpt4point_test}.
When evaluating image-based methods on ObjaNeRF--Text, we used random object views since Objaverse 3D models lack consistent orientation, making it impossible to identify standard front and back views as in ShapeNeRF--Text.

\model{}-13b consistently outperforms all other models across all metrics, often by large margins against runner-ups. Moreover, in most cases, the second best performing model is \model{}-7b, thus the same architecture with a smaller LLM. 
The difference in the quality of the caption generated by \model{} compared to the baselines is showcased by the qualitative results reported in  Figures~\ref{fig:qualitative_results_shapenerf_brief}, \ref{fig:qualitative_results_objanerf_pointllm}, and~\ref{fig:qualitative_results_objanerf_gpt4point} where the descriptions provided by \model{} are the most accurate.

\begin{table}[t]
    \centering
    \caption{\textbf{NeRF brief captioning on ShapeNeRF--Text.} \\Best results are in \textbf{bold}, runner-up is \underline{underlined}. \\(FV: front-view, BV: back-view, MV: multi-view)}
    \resizebox{\linewidth}{!}{%
        \begin{tabular}{lcccccc}
        \midrule
        \textbf{Model} & \textbf{Modality} & \textbf{S-BERT} & \textbf{SimCSE} & \textbf{BLEU-1} & \textbf{ROUGE-L} & \textbf{METEOR} \\
        \cmidrule(lr){1-1} \cmidrule(lr){2-2} \cmidrule(lr){3-7}
        GPT4Point-Opt-2.7b & Point cloud & 41.85 & 40.22 & 11.76 & 16.54 & 11.63 \\
        3D-LLM FlanT5-xl & Mesh + MV &  59.46 & 56.42 & 12.69 & 21.49 & 14.32 \\
        \llava{}-vicuna-7b & Image (FV) & \underline{59.85} & \underline{62.35} & 22.67 & \underline{23.24} & \underline{23.35} \\
        \llava{}-vicuna-7b & Image (BV) & 55.68 & 58.46 & 21.97 & 22.46 & 22.50 \\
        \llava{}-vicuna-7b & Image (MV) & 59.77 & 61.42 & \underline{23.16} & 22.68 & 23.01 \\
        PointLLM-7b & Point cloud & 49.59 & 48.84 & 16.74 & 17.92 & 14.56 \\
        %\model{}-7b & \nerf{} & 68.63 & 70.54 & 20.64 & 28.33 & 31.76 \\
        %\model{}-7b (ObjaNeRF, new training recipe) & \nerf{} & 43.97 & 44.26 & 14.36 & 18.93 & 12.93 \\
        \model{}-7b & \nerf{} & \textbf{74.94} & \textbf{76.41} & \textbf{42.73} & \textbf{43.64} & \textbf{41.95} \\
        
        \cmidrule(lr){1-1} \cmidrule(lr){2-2} \cmidrule(lr){3-7}
        BLIP-2 FlanT5-xxl & Image (FV) & 56.13 & 58.21 & 5.46  & 18.69 & 9.67 \\
        BLIP-2 FlanT5-xxl & Image (BV) & 52.48 & 54.05 & 5.67  & 18.20 & 9.50 \\
        \llava{}-vicuna-13b & Image (FV) & \underline{61.00} & \underline{61.16} & 14.30 & 20.00 & \underline{23.31} \\
        \llava{}-vicuna-13b & Image (BV) & 54.35 & 56.09 & 21.94 & 21.67 & 22.09 \\
        \llava{}-vicuna-13b & Image (MV) & 59.64 & 61.01 & \underline{22.84} & \underline{22.17} & 23.08 \\
        PointLLM-13b & Point cloud & 51.54 & 50.35 & 17.20 & 18.51 & 14.92 \\
        \model{}-13b & \nerf{} & \textbf{75.09} & \textbf{76.45} & \textbf{42.83} & \textbf{43.70} & \textbf{42.21} \\
        \bottomrule
        \end{tabular}}
        \label{tab:brief_frozen}
\end{table}

\begin{table}[t]
    \hfill
        \caption{\textbf{NeRF brief captioning on the HST dataset.} \\Best results are in \textbf{bold}, runner-up is \underline{underlined}. \\(FV: front-view, BV: back-view, MV: multi-view)}
        \resizebox{\linewidth}{!}{%
        \begin{tabular}{lcccccc}
        \midrule
        \textbf{Model} & \textbf{Modality} & \textbf{S-BERT} & \textbf{SimCSE} & \textbf{BLEU-1} & \textbf{ROUGE-L} & \textbf{METEOR} \\
        \cmidrule(lr){1-1} \cmidrule(lr){2-2} \cmidrule(lr){3-7}
        GPT4Point-Opt-2.7B & Point cloud & 43.15 & 42.22 & 12.02 & 18.73 & 13.69 \\
        3D-LLM FlanT5-xl & Mesh + MV &  \underline{56.07} & 52.13 & \underline{15.94} & \underline{20.71} & \underline{15.22} \\
        \llava{}-vicuna-7b & Image (FV)  & 54.31 & 56.28 & 10.08 & 14.71 & 14.53 \\
        \llava{}-vicuna-7b & Image (BV)     & 51.75 & 52.29 & 8.13 & 13.96 & 14.18 \\
        \llava{}-vicuna-7b & Image (MV)     & 54.15 & \underline{56.87} & 10.32 & 12.76 & 15.13 \\
        PointLLM-7b & Point cloud & 43.40 & 44.50 &  8.53 & 11.64 &  9.97 \\
        %\model{}-7b & \nerf{} & 59.20 & 61.66 &  9.47 & 14.94 & 17.06 \\
        %\model{}-7b (ObjaNeRF, new training recipe) & \nerf{} & 39.47 & 39.78 & 14.37 & 17.44 & 12.87 \\
        \model{}-7b & \nerf{} & \textbf{64.78} & \textbf{65.01} & \textbf{20.65} & \textbf{23.24} & \textbf{24.10} \\
        \cmidrule(lr){1-1} \cmidrule(lr){2-2} \cmidrule(lr){3-7}
        BLIP-2 FlanT5-xxl& Image (FV)& \underline{57.11} & \underline{59.43} &  8.21 & \underline{18.02} & 12.14 \\
        BLIP-2 FlanT5-xxl& Image (BV)   & 54.11 & 56.37 &  9.09 & 17.38 & 11.79 \\
        \llava{}-vicuna-13b& Image (FV) & 55.62 & 55.56 &  6.56 & 11.81 & 14.52 \\
        \llava{}-vicuna-13b& Image (BV)  & 50.00 & 50.79 & 9.39 & 12.76 & 14.46 \\
        \llava{}-vicuna-13b & Image (MV) & 54.25 & 55.56 & \underline{9.78} & 14.13 & \underline{14.99} \\
        PointLLM-13b & Point cloud & 45.41 & 46.39 & 9.57 & 12.38 & 11.92 \\
        \model{}-13b & \nerf{} & \textbf{65.66} & \textbf{66.09} & \textbf{20.80} & \textbf{24.28} & \textbf{25.90} \\
        \bottomrule
        \end{tabular}}
        \label{tab:brief_frozen_hst}
\end{table}
\begin{table}[t]
    \centering
    \caption{\textbf{NeRF brief captioning on ObjaNeRF--Text (PointLLM test set).} \\Best results are in \textbf{bold}, runner-up is \underline{underlined}. \\(RV: random view, MV: multi-view)}
    \resizebox{\linewidth}{!}{%
        \begin{tabular}{lcccccc}
        \midrule
        \textbf{Model} & \textbf{Modality} & \textbf{S-BERT} & \textbf{SimCSE} & \textbf{BLEU-1} & \textbf{ROUGE-L} & \textbf{METEOR} \\
        \cmidrule(lr){1-1} \cmidrule(lr){2-2} \cmidrule(lr){3-7}
        3D-LLM FlanT5-xl & Mesh + MV & 38.22 & 39.60 & 5.47 & 7.29 & 10.75 \\
        \llava{}-vicuna-7b & Image (RV) & 39.29 & 40.93 & 5.22 & 8.35 & 12.49 \\
        \llava{}-vicuna-7b & Image (MV) & \underline{40.15} & \underline{41.09} & \underline{5.62} & \underline{9.07} & \underline{13.22} \\
        PointLLM-7b & Point cloud & 38.81 & 40.11 & 5.55 & 8.39 & 11.24 \\
        %\model{}-7b & \nerf{} & 35.78 & 37.27 & 5.42 & 7.67 & 10.93 \\
        \model{}-7b & \nerf{} & \textbf{41.36} & \textbf{42.28} & \textbf{13.21} & \textbf{16.93} & \textbf{15.63} \\
        \cmidrule(lr){1-1} \cmidrule(lr){2-2} \cmidrule(lr){3-7}
        BLIP-2 FlanT5-xxl & Image (RV) & 35.48 & 35.19 & \underline{7.69}  & \underline{13.75} & 11.68 \\
        \llava{}-vicuna-13b & Image (RV) & 38.57 & 39.05 & 4.62 & 7.63 & 11.85 \\
        \llava{}-vicuna-13b & Image (MV) & \underline{41.01} & \underline{41.10} & 4.87 & 8.03 & \underline{12.35}\\
        PointLLM-13b & Point cloud & 39.64 & 40.63 & 5.94 & 8.43 & 11.63 \\
        \model{}-13b & \nerf{} & \textbf{42.08} & \textbf{42.40} & \textbf{13.86} & \textbf{17.51} & \textbf{16.18} \\
        \bottomrule
        \end{tabular}}
        \label{tab:objanerf_pointllm_test}
\end{table}

\begin{table}[t]
    \centering
    \caption{\textbf{NeRF brief captioning on ObjaNeRF--Text (GPT4Point test set).} \\Best results are in \textbf{bold}, runner-up is \underline{underlined}. \\(RV: random view, MV: multi-view)}
    \resizebox{\linewidth}{!}{%
        \begin{tabular}{lcccccc}
        \midrule
        \textbf{Model} & \textbf{Modality} & \textbf{S-BERT} & \textbf{SimCSE} & \textbf{BLEU-1} & \textbf{ROUGE-L} & \textbf{METEOR} \\
        \cmidrule(lr){1-1} \cmidrule(lr){2-2} \cmidrule(lr){3-7}
        GPT4Point-Opt-2.7b & Point cloud & 34.42 & 31.89 & 8.76 & \underline{13.62} & 13.55 \\
        3D-LLM FlanT5-xl & Mesh + MV & 41.26 & 37.74 & \underline{11.08} & 12.26 & 15.58 \\
        \llava{}-vicuna-7b & Image (RV) & 41.59 & 42.38 & 7.24 & 11.74 & 16.01 \\
        \llava{}-vicuna-7b & Image (RV) & \underline{42.34} & \underline{43.21} & 8.59 & 11.94 & \underline{18.52} \\
        %\model{}-7b & \nerf{} & 33.08 & 31.87 & 6.41 & 8.89 & 12.37 \\
        %\model{}-7b (ObjaNeRF, new training recipe) & \nerf{} & 44.38 & 43.25 & 21.53 & 26.40 & 23.27 \\
        \model{}-7b & \nerf{} & \textbf{43.73} & \textbf{43.09} & \textbf{20.22} & \textbf{25.15} & \textbf{22.21} \\
        \cmidrule(lr){1-1} \cmidrule(lr){2-2} \cmidrule(lr){3-7}
        BLIP-2 FlanT5-xxl & Image (RV) & 37.24 & 37.32 & \underline{10.63} & \underline{16.87} & 14.69 \\
        \llava{}-vicuna-13b & Image (RV) & 42.08 & 41.04 & 6.53 & 10.57 & 15.67 \\
        \llava{}-vicuna-13b & Image (MV) & \underline{44.15} & \underline{43.19} & 6.75 & 10.86 & \underline{16.10} \\
        \model{}-13b & \nerf{} & \textbf{44.26} & \textbf{43.75} & \textbf{20.61} & \textbf{25.62} & \textbf{22.36} \\
        \bottomrule
        \end{tabular}}
        \label{tab:objanerf_gpt4point_test}
\end{table}

A clear trend in the tables and qualitative results is that image-based models tend to perform better than models processing point clouds. 
This is likely due to the larger amount of data used during training of the modality encoder, i.e. millions of images versus hundreds of thousands of shapes, which enhances their generalization ability, as well as the capability of images to capture more details than point clouds at the input resolutions required by image-based MLLMs versus 3D MLLMs. 
Nonetheless, our method, which operates on NeRFs, benefits from a holistic view of the object and provides the most accurate descriptions. 
Remarkably, in \model{}, all the necessary information for this language task can be extracted from a single global embedding obtained by directly processing the NeRF weights. 
Comparing the results of image-based MLLMs when processing front versus back views, we can see that the vantage point has a non-negligible effect on the performance of such baselines, with SentenceBERT and SimCSE metrics diminishing by about $4$ points in all baselines.
In a dataset without canonical poses for objects, this would be a relevant limitation that processing NeRF weights seamlessly sidesteps. 
Finally, we observe that the multi-view setup of \llava{} provides better performance to the single-view counterpart.
Regarding LLM size scaling, our analysis reveals only marginal performance improvements with larger models in brief captioning tasks. 
On ShapeNeRF-Text, \llava{} shows a minimal improvement in S-BERT scores from $59.85$ to $61.00$ when using the 13B parameter model, while other metrics actually deteriorate. Similarly, PointLLM exhibits modest gains of approximately $1$ point across metrics, and LLaNA demonstrates even smaller improvements. 
This trend is consistent across HST and ObjaNeRF-Text datasets. 
These findings suggest that the ability to perform NeRF-language tasks is not strongly correlated with LLM size.
Unlike traditional NLP tasks, where larger models generally lead to significant performance boosts due to their enhanced language understanding and generation capabilities, NeRF-based captioning appears to depend more on the model’s ability to process and integrate 3D and visual information from the input \nerf{}. 
Given the substantial computational costs of training and deploying larger models, the minimal performance gains observed in brief captioning tasks may not justify the increased resource demands.
Another key observation is that the quality of the input encoding and the processing pipeline, which turns \nerf{} representations into LLM-compatible features, have a greater impact on performance than increasing the size of the underlying LLM.
For example, despite using LLMs of similar sizes (2.7B for GPT4Point and 3B for 3D-LLM), these models exhibit very different performance levels. One possible explanation is that these tasks require precise spatial and geometric reasoning, which may not inherently improve with a larger LLM. 
Models like 3D-LLM, which incorporate multi-view processing and colored meshes, likely benefit more from their specialized architecture than from parameter scaling.

\begin{figure*}[t]
    \centering
    \includegraphics[width=\linewidth]{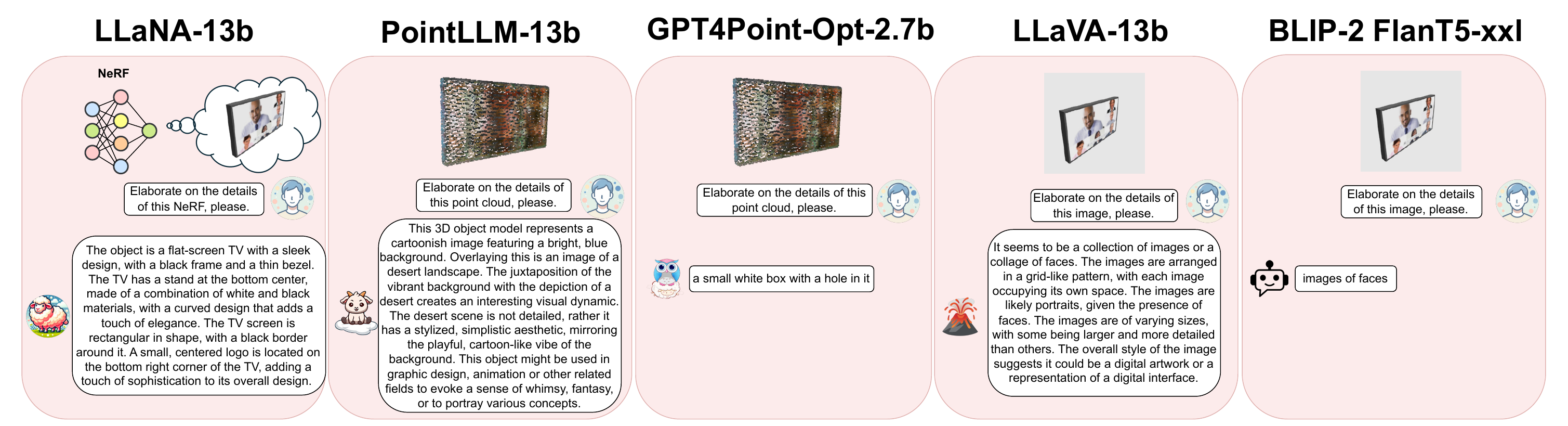}
    \caption{\textbf{Qualitative results on ShapeNeRF--Text detailed descriptions.} From top to bottom: brief and detailed descriptions, single-round Q\&A}
    \label{fig:qualitative_results_shapenerf_detailed}
\end{figure*}

\begin{table}[t]
    \caption{\textbf{NeRF detailed captioning on ShapeNeRF--Text.}\\Best results are in \textbf{bold}, runner-up is \underline{underlined}. \\(FV: front-view, BV: back-view, MV: multi-view)}
    \centering
    \resizebox{\linewidth}{!}{
    \begin{tabular}{lcccccc}
        \midrule
        \textbf{Model} & \textbf{Modality} & \textbf{S-BERT} & \textbf{SimCSE} & \textbf{BLEU-1} & \textbf{ROUGE-L} & \textbf{METEOR} \\
        \cmidrule(lr){1-1} \cmidrule(lr){2-2} \cmidrule(lr){3-7}
        GPT4Point-Opt-2.7b & Point cloud  & 42.44 & 38.33 & 3.72 & 9.21 & 5.13 \\
        3D-LLM FlanT5-xl & Mesh + MV & 58.00 & 53.91 & 1.58 & 14.40 & 5.28 \\
        \llava{}-vicuna-7b& Image (FV)  & 57.55 & 57.68 & 14.99 & 22.82 & 14.36 \\
        \llava{}-vicuna-7b& Image (BV) & 53.11 & 54.46 & 14.73 & 22.47 &	14.05 \\
        \llava{}-vicuna-7b& Image (MV) & 55.26 & \underline{58.46} & \underline{15.07} & \underline{24.05} & \underline{14.85} \\
        PointLLM-7b & Point cloud & \underline{59.02} & 58.30 & 10.28 & 19.26 & 10.55 \\
        %\model{}-7b & \nerf{} & 77.43 & 79.81 & 41.32 & 36.18 & 32.39 \\
        %\model{}-7b (ObjaNeRF, new training recipe) & \nerf{} & 50.89 & 49.25 & 10.76 & 18.93 & 10.66 \\
        \model{}-7b & \nerf{} & \textbf{75.25} & \textbf{77.42} & \textbf{19.57} & \textbf{32.96} & \textbf{20.45} \\
        \cmidrule(lr){1-1} \cmidrule(lr){2-2} \cmidrule(lr){3-7}
        BLIP-2 FlanT5-xxl& Image (FV) & 41.27 & 40.69 &  0.18 &  7.83 &  2.60 \\
        BLIP-2 FlanT5-xxl& Image (BV) & 38.49 & 37.89 &  0.19 &  7.72 &  2.58 \\
        \llava{}-vicuna-13b& Image (FV) & 59.08 & 58.87 & \textbf{23.63} & 23.55 & \textbf{22.55} \\
        \llava{}-vicuna-13b& Image (BV) & 50.09 & 50.33 & 13.77 & 21.36 & 13.18 \\
        \llava{}-vicuna-13b & Image (MV) & \underline{60.21} & \underline{59.51} & 15.07 & \underline{32.16} & 14.64\\
        PointLLM-13b & Point cloud & 59.64 & 58.55 & 10.52 & 19.44 & 10.83 \\
        \model{}-13b & \nerf{} & \textbf{75.51} & \textbf{77.63} & \underline{19.87} & \textbf{32.93} & \underline{20.46} \\
        
        \bottomrule
    \end{tabular}}
    \label{tab:detailed_frozen}
\end{table}

\paragraph{Detailed captioning}
The results for the detailed captioning task are presented in \cref{tab:detailed_frozen}. \model{} demonstrates superior performance compared to all other models, showing significant improvements in data-driven metrics such as Sentence-BERT and SimCSE. For traditional metrics like BLEU-1, ROUGE-L, and METEOR, our model achieves comparable results to \llava{}. 
3D-LLM \cite{3dllm}, processing multi-view images and colored meshes, performs well on the Sentence-BERT metric, whereas all other metrics show poor results. Interestingly, the point-based model PointLLM \cite{pointllm} performs similarly to the image based one, \llava{}~\cite{llava}. Considering the Sentence-BERT metric, \model{}-13b achieves $75.51$, notably $15.87$ points more than PointLLM and $15.30$ points more than \llava{}-13b multi-view setup. 
These substantial performance gaps suggest that, while individual or aggregated images may be sufficient for brief descriptions, they may lack all the details needed to provide a comprehensive description. 
Moreover, the dependency of the output quality on the selected vantage points remains strong, as proven by the varying performance achieved by \llava{} across front-view, back-view, and multi-view scenarios. 
In contrast, the NeRF weights contain detailed and complete information about the object, which is fundamental for more granular description tasks, with the additional advantage of not requiring tuning such hyperparameters.
\begin{table}[t]
    \centering
    \caption{\textbf{NeRF single-round Q\&A on ShapeNeRF--Text.}\\Best results are in \textbf{bold}, runner-up is \underline{underlined}. \\(FV: front-view, BV: back-view, MV: multi-view)}
        \resizebox{\linewidth}{!}{%
        \begin{tabular}{lcccccc}
        \midrule
        \textbf{Model} & \textbf{Modality} & \textbf{S-BERT} & \textbf{SimCSE} & \textbf{BLEU-1} & \textbf{ROUGE-L} & \textbf{METEOR} \\
        \cmidrule(lr){1-1} \cmidrule(lr){2-2} \cmidrule(lr){3-7}
        GPT4Point-Opt-2.7b & Point cloud & 27.62 & 31.41 & 6.26 & 9.38 & 5.41 \\
        3D-LLM FlanT5-xl & Mesh + MV & 69.62 & 67.55 & 32.19 & 40.95 & 35.83\\
        \llava{}-vicuna-7b& Image (FV) & 71.79 & 71.96 & 25.79 & 34.04 & 34.86 \\
        \llava{}-vicuna-7b& Image (BV) & 70.88 & 70.93 & 25.17 & 33.30 & 34.22 \\
         \llava{}-vicuna-7b& Image (MV) & 72.26 & 70.67 & 24.09 & 34.67 & 35.64 \\
        PointLLM-7b & Point cloud & \underline{74.70} & \underline{74.40} & \underline{36.81} & \underline{44.41} & \underline{39.76} \\
        %\model{}-7b & \nerf{} & 81.03 & 81.56 & 46.16 & 53.17 & 50.15 \\
        %\model{}-7b (ObjaNeRF, new training recipe) & \nerf{} & 73.47 & 72.85 & 37.42 & 45.26 & 40.57 \\
        \model{}-7b & \nerf{} & \textbf{81.03} & \textbf{81.61} & \textbf{45.98} & \textbf{53.27} & \textbf{49.97} \\
        \cmidrule(lr){1-1} \cmidrule(lr){2-2} \cmidrule(lr){3-7}
        BLIP-2 FlanT5-xxl& Image (FV) & 45.20 & 47.92 & 11.50 & 20.16 & 13.49 \\
        BLIP-2 FlanT5-xxl& Image (BV) & 45.06 & 47.66 & 11.50 & 19.98 & 13.44 \\
        \llava{}-vicuna-13b & Image (FV) & 71.61 & 70.98 & 20.19 & 30.42 & 32.53 \\
        \llava{}-vicuna-13b & Image (BV) & 68.25 & 69.06 & 20.03 & 29.84 & 32.27 \\
        \llava{}-vicuna-13b & Image (MV) & 71.84 & 71.16 & 20.04 & 30.20 & 33.46 \\
        PointLLM-13b & Point cloud & \underline{74.65} & \underline{74.18} & \underline{37.16} & \underline{44.86} & \underline{40.13} \\
        \model{}-13b & \nerf{} & \textbf{81.05} & \textbf{81.60} & \textbf{46.08} & \textbf{53.44} & \textbf{49.98} \\   
        
        \bottomrule
        \end{tabular}}
    \label{tab:qa_frozen}
\end{table}
The ability of NeRF to capture holistic information about the object is also shown in \cref{fig:qualitative_results_shapenerf_detailed}, where only the direct processing of NeRF weights lets LLaNA understand that the object is a TV. 
PointLLM and LLaVA provide detailed but wrong descriptions, likely because of the need to extract the intermediate discrete representation as a point cloud or an image, losing information. 
Indeed, in both cases, it is hard even for a human observer to provide the right description from the intermediate modalities shown in the figure.
When comparing the different versions of our model, the 13B variant slightly outperforms its 7B counterpart, a pattern consistently observed across other models in the table, such as PointLLM and \llava{}. 
Specifically, the relative improvements from 7B to 13B versions on Sentence-BERT and SimCSE are: $0.62$ and $0.25$ for PointLLM, $1.53$ and $1.19$ for \llava{}, $0.26$ and $0.21$ for \model{}. 
This limited performance gain suggests that the architectural improvements and training strategies employed in these models may be more crucial for performance than simply scaling up model size, as also observed with the results obtained in the brief captioning task.
Regarding the training strategies of these models, an interesting observation can be made. 
Overall, the highest performance on \nerf{} detailed captioning is achieved by \model{}, \llava{}, and PointLLM; notably, the only models incorporating LLM fine-tuning in their training protocol. 
This pattern strongly suggests that fine-tuning the LLM plays a crucial role in enhancing the quality of the generated descriptions. 
While this fine-tuning stage appears less critical for generating brief captions, it significantly impacts the quality of longer, detailed descriptions. 
This can be attributed to the ability of the fine-tuned LLM to adapt its pre-trained language understanding capabilities to the specific characteristics and vocabulary of \nerf{}-based object descriptions.

\begin{figure*}[t]
    \centering
    \includegraphics[width=\linewidth]{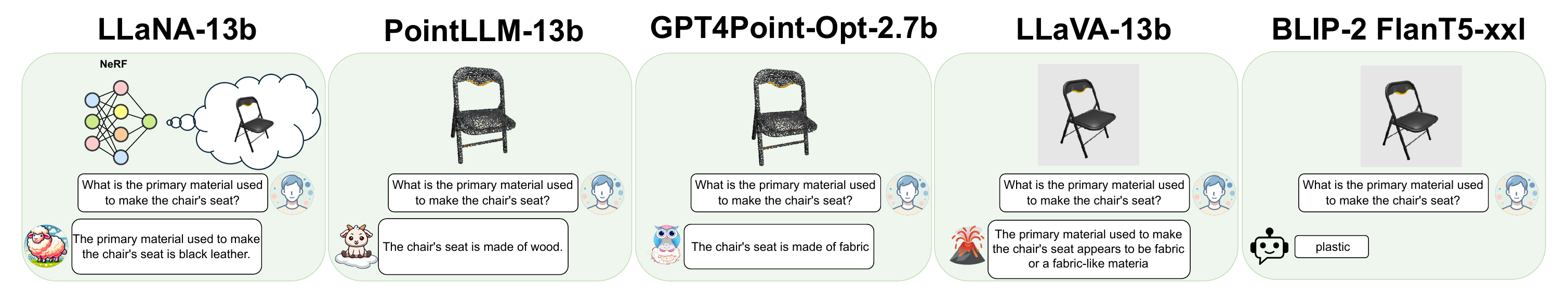}
    \caption{\textbf{Qualitative results on ShapeNeRF--Text single-round Q\&A.}}
    \label{fig:qualitative_results_shapenerf_qa}
\end{figure*}

\begin{figure}[t]
         \begin{center}
         \includegraphics[width=0.8\linewidth]{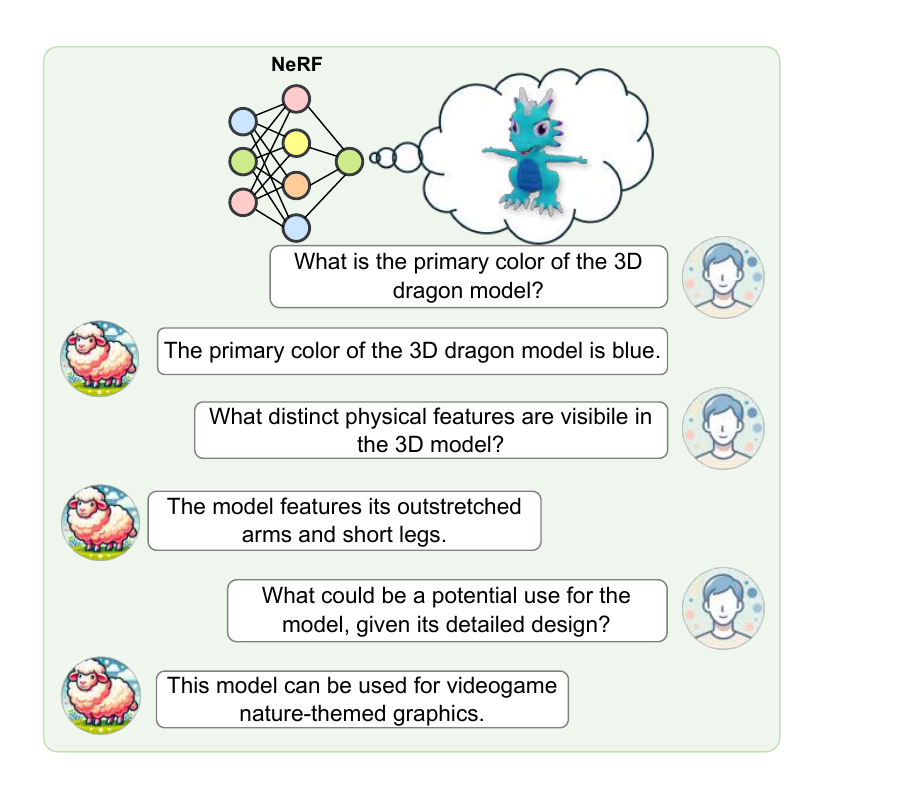}
         \captionof{figure}{\textbf{NeRF multi-round Q\&A example from ObjaNeRF--Text.}}
         \label{fig:multi-round}
         \end{center}
 \end{figure}

\paragraph{single-round Q\&A}
In the single-round Q\&A experiment, we test the ability of the assistants to provide accurate answers to specific questions about the object.
We prompt the models with the \nerf{}, or the image/cloud extracted from it, followed by one of the questions in the single-round Q\&A annotations associated with the \nerf{}.
We then collect the answer generated by the model and compare it against the ground-truth answer with the selected metrics. 
Results are reported in \cref{tab:qa_frozen}.
Interestingly, PointLLM~\cite{pointllm} performs better than \llava{}~\cite{llava} in this task, likely because it has been specifically trained to answer detailed questions about objects represented as point clouds.
Nevertheless, \model{} maintains its position as the top-performing method across all metrics by substantial margins, mirroring our findings from the brief and detailed captioning tasks.
Using the 13B LLAMA backbone, the performance gaps between \model{} and the second-best model, PointLLM, are large: $6.40$ for Sentence-BERT, $7.42$ for SimCSE, $8.92$ for BLEU-1, $8.58$ for ROUGE-L, and $9.85$ for METEOR. 
Notably, these margins remain consistently large even when using the 7B LLAMA backbone. 
The performance advantage of \model{} shows that our meta-encoder and projector architecture are capable of effectively extracting fine-grained information from the \nerf{} representation, even if they are processing directly \nerf{} weights. 
Remarkably, the amount of information they can extract lets \model{} answer more precisely than when images or point clouds are extracted from the NeRF. 
Indeed, as shown in \cref{fig:qualitative_results_shapenerf_qa} which reports a qualitative example from ShapeNeRF--Text, the only assistant able to answer correctly to a precise question about the material of the chair is \model{}. 
Finally, another qualitative result confirming the ability of \model{} to provide high-quality answers to specific questions, in this case in a multi-round Q\&A experiment where a human user asks questions on a NeRF from the test set of ObjaNeRF--Text, is reported in \cref{fig:multi-round}.
Similar to our findings with brief and detailed descriptions, the Q\&A results show minimal performance difference between \model{}-13B and \model{}-7B, further reinforcing that MLLM performance on these tasks is not strongly dependent on the size of the underlying language model. 
Furthermore, the consistently superior performance of \model{}, \llava{}, and PointLLM across both tasks underscores the critical role of LLM finetuning in developing models that can effectively describe and answer questions regarding 3D objects.

\begin{table}[t]
    \caption{\textbf{Zero-Shot NeRF Classification.}\\Best results are in \textbf{bold}, runner-up is \underline{underlined}. \\(FV: front-view, BV: back-view, MV: multi-view)}
    \centering
    \resizebox{0.8\linewidth}{!}{
    \begin{tabular}{lcc}
        \midrule
        \textbf{Model} & \textbf{Modality} & \textbf{Accuracy (\%)} \\
        \cmidrule(lr){1-1} \cmidrule(lr){2-2} \cmidrule(lr){3-3}
        GPT4Point-Opt-2.7b & Point cloud & 41.93 \\
        3D-LLM FlanT5-xl & Mesh + MV & 60.55 \\
        \llava{}-vicuna-7b & Image (FV) & 60.25 \\
        \llava{}-vicuna-7b & Image (BV) & 57.00 \\
        \llava{}-vicuna-7b & Image (MV) & \underline{67.25}\\
        PointLLM-7b & Point cloud & 50.14 \\
        %\model{}-7b & \nerf{} & 67.14 \\
        %\model{}-7b (ObjaNeRF, new training recipe) & \nerf{} & 38.51\\
        \model{}-7b & \nerf{} & \textbf{67.56} \\
        \cmidrule(lr){1-1} \cmidrule(lr){2-2} \cmidrule(lr){3-3}
        BLIP-2 FlanT5-xxl & Image (FV) & 63.67 \\
        BLIP-2 FlanT5-xxl & Image (BV) & 61.47 \\
        \llava{}-vicuna-13b & Image (FV) & 66.13 \\
        \llava{}-vicuna-13b & Image (BV) & 63.90 \\
        \llava{}-vicuna-13b & Image (MV) & \textbf{73.45} \\
        PointLLM-13b & Point cloud & 48.72 \\
        \model{}-13b & \nerf{} & \underline{69.27} \\    
        \bottomrule
        \end{tabular}}
    \label{tab:classification_frozen}
\end{table}

\paragraph{Zero-shot classification}
We compare assistants on the task of zero-shot classification. 
We query the models with the sentence \emph{``What is the class of the NeRF/image/cloud? Choose among these: [ShapeNet classes]''} where \emph{[ShapeNet classes]} are the 10 ShapeNet classes available in ShapeNeRF--Text. 
We consider the answer correct only if the ground truth class appears in the response.
We report results in \cref{tab:classification_frozen} on ShapeNeRF--Text.
Using multiple views boosts the zero-shot classification performance of \llava{}-13b, which turns out to be the best model for this task, followed by \model{}-13b. 
Similar to the brief captioning task, image-based models tend to outperform point cloud-based models on this classification task. 
This performance pattern aligns with the requirements of these tasks. 
For brief captioning and classification, which primarily require high-level understanding and concise outputs, image-based models excel by leveraging visual features directly from 2D views, where the nature of the object and its appearance are readily accessible.
However, the pattern reverses for detailed captioning and Q\&A tasks, where geometric precision and spatial understanding become crucial. These tasks often require reasoning about specific object parts, their relationships, and fine-grained spatial details - information that is inherently preserved in point cloud representations. While image-based models might struggle with occlusions and loss of information due to the chosen vantage points, point cloud-based approaches can directly reason about the complete 3D geometry, leading to more accurate and detailed responses.

\begin{table}[t]
    \begin{minipage}[t]{0.48\textwidth}
        \centering
        \caption{\textbf{NeRF brief captioning on ShapeNeRF--Text.} All methods trained on ShapeNeRF--Text training set. Best results are in \textbf{bold}, runner-up is \underline{underlined}. \\(FV: front-view, MV: multi-view)}
        \resizebox{\linewidth}{!}{
            \begin{tabular}{lcccccc}
            \midrule
            \textbf{Model} & \textbf{Modality} & \textbf{S-BERT} & \textbf{SimCSE} & \textbf{BLEU-1} & \textbf{ROUGE-L} & 
            \textbf{METEOR} \\
            \cmidrule(lr){1-1} \cmidrule(lr){2-2} \cmidrule(lr){3-7}
            GPT4Point-Opt-2.7b & Point cloud & 37.96 & 39.00 & 21.33 & 22.29 & \underline{24.88} \\
            PointLLM-7b & Point cloud & \underline{55.48} & \underline{57.28} & \underline{21.67} &   \underline{25.84} & 24.54 \\
            %\model{}-7b & \nerf{} & 68.63 & 70.54 & 20.64 & 28.33 & 31.76 \\
            \model{}-7b & \nerf{} & \textbf{76.04} & \textbf{77.13} & 
            \textbf{45.00} & \textbf{45.87} & \textbf{44.52} \\
            \cmidrule(lr){1-1} \cmidrule(lr){2-2} \cmidrule(lr){3-7}
            \llava{}-vicuna-13b & Image (FV) & 42.86 & 43.22 & 15.56 & 13.74 & 15.27 \\
            \llava{}-vicuna-13b & Image (MV) & 44.29 & 41.85 & 16.67 & 16.26 & 17.58 \\
            \bottomrule
            \end{tabular}}
        \label{tab:brief_trained}
    \end{minipage}
    \hfill
    \vspace{\baselineskip}
    \begin{minipage}[t]{0.48\textwidth}
        \centering
        \caption{\textbf{NeRF brief captioning on the HST dataset.} All methods trained on ShapeNeRF--Text training set. Best results are in \textbf{bold}, runner-up is \underline{underlined}. \\(FV: front-view, MV: multi-view)}
        \resizebox{\linewidth}{!}{
            \begin{tabular}{lcccccc}
            \midrule
            \textbf{Model} & \textbf{Modality} & \textbf{S-BERT} & \textbf{SimCSE} & \textbf{BLEU-1} & \textbf{ROUGE-L} & \textbf{METEOR} \\
            \cmidrule(lr){1-1} \cmidrule(lr){2-2} \cmidrule(lr){3-7}
            GPT4Point-Opt-2.7B & Point cloud & 30.50 & 31.08 & 8.12 & 12.35 & 11.62 \\
            PointLLM-7b & Point cloud & \underline{44.65} & \underline{44.68} & 8.91 & 12.33 &  \underline{12.64} \\
            %\model{}-7b & \nerf{} & \underline{55.62} & \underline{55.56} & 6.56 & 11.81 & \underline{14.52}  \\
            \model{}-7b & \nerf{} & \textbf{65.42} & \textbf{66.44} & \textbf{19.63} & \textbf{23.39} & \textbf{23.46} \\
            \cmidrule(lr){1-1} \cmidrule(lr){2-2} \cmidrule(lr){3-7}
            \llava{}-vicuna-13b & Image (FV) & 33.79 & 42.66 & 10.28 & \underline{13.22} & 12.19 \\   
            \llava{}-vicuna-13b & Image (MV) & 34.45 & 42.54 & \underline{11.04} & 12.36 & 11.48 \\ 
            \bottomrule
            \end{tabular}}
        \label{tab:brief_trained_hst}
    \end{minipage}
\end{table}

\subsection{Ablation study on training data }

In this section, we run an ablation study where we train from scratch some methods on the same training set, i.e. ShapeNeRF--Text. This is done to assess the influence of the training set on the results. 
Due to the considerable computational resources required for training these models, we evaluated a subset of baselines using their official training code. 
Accordingly, we followed their protocol, which, for all methods, keeps the modality-specific encoder frozen and trains an adaptor. 
In PointLLM and \llava{}, the LLM is finetuned during an additional training stage. 

Tables~\ref{tab:brief_trained}, \ref{tab:brief_trained_hst}, \ref{tab:detailed_trained}, \ref{tab:qa_trained} and \ref{tab:classification_trained} report results for both \model{} and the baselines trained solely on ShapeNeRF--Text. 
We notice that these baselines exhibit different behaviors to their pre-trained counterparts, with \llava{} performing significantly worse and PointLLM showing clear improvements. As for GPT4Point, we observe greater variability across metrics; however, overall, it shows no significant benefit from training on ShapeNeRF--Text. Also in this scenario, \model{}
yields the best performance compared to all baselines.

\begin{table}[t]
    \begin{minipage}[t]{0.48\textwidth}
        \centering
        \caption{\textbf{NeRF detailed captioning on ShapeNeRF--Text.} All methods trained on ShapeNeRF--Text training set. Best results are in \textbf{bold}, runner-up is \underline{underlined}. \\(FV: front-view, MV: multi-view)}
        \resizebox{\linewidth}{!}{
            \begin{tabular}{lcccccc}
        \midrule
        \textbf{Model} & \textbf{Modality} & \textbf{S-BERT} & \textbf{SimCSE} & \textbf{BLEU-1} & \textbf{ROUGE-L} & \textbf{METEOR} \\
        \cmidrule(lr){1-1} \cmidrule(lr){2-2} \cmidrule(lr){3-7}
        GPT4Point-Opt-2.7b & Point cloud & 41.33 & 40.52 & 14.48 & 19.15 & 13.80 \\
        PointLLM-7b & Point cloud & \underline{67.30} & \underline{59.56} & \underline{15.39} & 21.42 & 11.37 \\
        %\model{}-7b & \nerf{} & \textbf{77.43} & \textbf{79.81} & \textbf{41.32} & \textbf{36.18} & \textbf{32.39} \\
        \model{}-7b & \nerf{} & \textbf{75.83} & \textbf{77.92} & \textbf{20.22} & \textbf{33.80} & \textbf{21.20} \\
        \cmidrule(lr){1-1} \cmidrule(lr){2-2} \cmidrule(lr){3-7}
        \llava{}-vicuna-13b & Image (FV) & 44.69 & 42.31 & 10.08 & 23.46 & 12.70 \\   
        \llava{}-vicuna-13b & Image (MV) & 45.72 & 42.49 & 12.17 & \underline{23.48} & \underline{14.25} \\
        \bottomrule
        \end{tabular}}
    \label{tab:detailed_trained}
    \end{minipage}
    \hfill
    \vspace{\baselineskip}
    \begin{minipage}[t]{0.48\textwidth}
        \centering
        \caption{\textbf{NeRF single-round Q\&A on ShapeNeRF--Text} All methods trained on ShapeNeRF--Text training set. Best results are in \textbf{bold}, runner-up is \underline{underlined}. \\(FV: front-view, MV: multi-view)}
        \resizebox{\linewidth}{!}{
            \begin{tabular}{lcccccc}
            \midrule
            \textbf{Model} & \textbf{Modality} & \textbf{S-BERT} & \textbf{SimCSE} & \textbf{BLEU-1} & \textbf{ROUGE-L} & \textbf{METEOR} \\
            \cmidrule(lr){1-1} \cmidrule(lr){2-2} \cmidrule(lr){3-7}
            GPT4Point-Opt-2.7b & Point cloud & 22.22 & 28.66 & 8.76 & 13.46 & 14.19 \\
            PointLLM-7b & Point cloud & \underline{79.24} & \underline{80.38} & \underline{46.00} & \underline{52.60} & \underline{42.36} \\
            %\model{}-7b & \nerf{} & \underline{81.03} & \underline{81.56} & \underline{46.16} & \underline{53.17} & \underline{50.15} \\
            \model{}-7b & \nerf{} & \textbf{81.94} & \textbf{82.43} & \textbf{47.73} & \textbf{54.97} & \textbf{51.78} \\
            \cmidrule(lr){1-1} \cmidrule(lr){2-2} \cmidrule(lr){3-7}
            \llava{}-vicuna-13b & Image (FV) & 56.29 & 62.36 & 26.87 & 29.55 & 30.49 \\    
            \llava{}-vicuna-13b & Image (MV) & 56.42 & 64.47 & 25.41 & 30.09 & 32.04 \\     
            \bottomrule
            \end{tabular}}
        \label{tab:qa_trained}
    \end{minipage}
\end{table}
\begin{table}[t]
    \caption{\textbf{Zero-shot NeRF classification on ShapeNeRF--Text.} All methods trained on ShapeNeRF--Text training set. Best results are in \textbf{bold}, runner-up is \underline{underlined}. \\(FV: front-view, MV: multi-view)}
    \centering
    %\resizebox{\linewidth}{!}{%
    \begin{tabular}{lcc}
        \midrule
        \textbf{Model} & \textbf{Modality} & \textbf{Accuracy (\%)} \\
        \cmidrule(lr){1-1} \cmidrule(lr){2-2} \cmidrule(lr){3-3}
        GPT4Point-Opt-2.7b & Point cloud & 26.30 \\
        PointLLM-7b & Point cloud & \underline{49.69} \\
        \model{}-7b & \nerf{} & \textbf{69.06} \\
        \cmidrule(lr){1-1} \cmidrule(lr){2-2} \cmidrule(lr){3-3}
        \llava{}-vicuna-13b & Image (FV) & 36.49 \\
        \llava{}-vicuna-13b & Image (FV) & 39.27 \\
        \bottomrule
    \end{tabular}
    %}
    \label{tab:classification_trained}
\end{table}

%\subsection{Ablation study on training stages of \model{}}
%\input{tables/training_stages}
%In order to assess the effectiveness of our training protocol, we evaluated the performance of \model{} at the end of both training stages. Results are reported in \cref{tab:training_stages}.

\subsection{Is the LLM all you need?}
\label{sec:language_only}
Finally, we investigate how LLAMA 2, the LLM on which \model{} relies, performs on \nerf{}-language tasks. 
For this evaluation protocol, the LLM, finetuned during the second training stage of \model{}, is provided with questions belonging to our datasets, requiring it to generate correct answers without access to \nerf{} data. 
Consequently, its predictions rely solely on textual patterns present in the training set. 
These experiments offer valuable insights into the language annotations in ShapeNeRF--Text and ObjaNeRF--Text, as well as the impact of different LLM sizes. Results are shown in \cref{tab:language_only}.
Comparing this table with the ones reporting the results of \model{}, we observe a significant performance gap between \model{} and LLAMA. 
For instance, using the Sentence-BERT metric, \model{} achieves scores of $75.09$ and $75.51$ on the brief and detailed captioning tasks of ShapeNeRF--Text, respectively, while LLAMA-13B attains only $29.29$ and $40.20$. 
This corresponds to performance drops of approximately $61\%$ and $47\%$. 
Similarly, for the brief captioning task on ObjaNeRF--Text, the performance drop is around $45\%$ on the PointLLM test set and $47\%$ on the GPT4Point test set. 
These substantial gaps indicate that ShapeNeRF--Text and ObjaNeRF--Text provide language tasks that require access to 3D object information. 
Without the information from the \nerf{} token, the LLM is not able to provide correct descriptions. Therefore, our datasets can be used as reliable benchmarks for evaluating NeRF-language tasks.
With regards to the single-round Q\&A annotations, LLAMA-2 can answer correctly to a large set of questions leading to a limited performance gap with \model{}: from $81.05$ to $76.85$, corresponding to a relative decrease of $5\%$. 
This relatively small difference can be attributed to the nature of the questions themselves: many of them rely on common sense and general object knowledge rather than specific 3D object information. 
For instance, questions like \emph{How can the filing cabinet be used to organize office documents?}, \emph{What is a suitable use for this table?} can be answered without detailed information from the corresponding 3D objects. 
These types of questions were deliberately included to preserve the strong reasoning capabilities of LLAMA while creating more natural and comprehensive conversations.
An additional noteworthy finding concerns the relationship between LLM size and performance on NeRF-language tasks. 
As discussed previously, using a larger LLM does not significantly improve the performance of \model{}. 
This pattern is also evident in \cref{tab:language_only}, where LLAMA-7b and LLAMA-13b provide very similar results on ShapeNeRF--Text and ObjaNeRF--Text. 
This suggests that the ability of the model to process and comprehend 3D object inputs - through its pre-trained encoder and projection layers - is more important than the size of the LLM.
\begin{table}[t]
    \centering
    \caption{\textbf{Language-only baselines NeRF captioning and NeRF single-round Q\&A.} 
    \\(Shape: ShapeNeRF--Text, Obja-P: ObjaNeRF--Text PointLLM test set, Obja-G: ObjaNeRF--Text GPT4Point test set)}
    \resizebox{\linewidth}{!}{%
        \begin{tabular}{lllccccc}
        \midrule
        \textbf{Model} & \textbf{Task} & \textbf{Dataset} & \textbf{SBERT} & \textbf{SimCSE} & \textbf{BLEU-1} & \textbf{ROUGE-L} & \textbf{METEOR} \\
        \cmidrule(lr){1-1} \cmidrule(lr){2-2} \cmidrule(lr){3-3} \cmidrule(lr){4-8} 
        LLaMA-7b  & brief  & Shape & 31.08 & 27.44 & 22.13 & 19.39 & 17.44 \\
        LLaMA-13b & brief  & Shape & 29.29 & 24.69 & 16.95 & 18.93 & 13.85 \\
        LLaMA-7b  & brief  & HST & 26.05 & 21.48 & 13.85 & 14.74 & 10.48 \\
        LLaMA-13b & brief  & HST & 26.83 & 21.44 & 13.19 & 15.30 & 10.12 \\
        LLaMA-7b  & brief  & Obja-P & 28.20 & 29.33 & 10.37 & 13.59 & 12.95 \\
        LLaMA-13b & brief  & Obja-P & 23.19 & 21.29 & 10.25 & 13.45 & 10.91 \\
        LLaMA-7b  & brief  & Obja-G & 23.82 & 21.70 & 13.21 & 16.79 & 13.50 \\
        LLaMA-13b & brief  & Obja-G & 23.43 & 19.54 & 14.18 & 17.40 & 13.81 \\
        \cmidrule(lr){1-1} \cmidrule(lr){2-2} \cmidrule(lr){3-3} \cmidrule(lr){4-8} 
        LLaMA-7b  & detailed & Shape & 37.51 & 32.47 & 9.66  & 16.67 & 9.67 \\
        LLaMA-13b & detailed & Shape & 40.20 & 35.80 & 11.35 & 18.82 & 11.29 \\
        \cmidrule(lr){1-1} \cmidrule(lr){2-2} \cmidrule(lr){3-3} \cmidrule(lr){4-8} 
        LLaMA-7b  & Q\&A  & Shape & 76.89 & 76.61 & 42.01 & 49.74 & 45.69 \\
        LLaMA-13b & Q\&A  & Shape & 76.85 & 76.79 & 42.30 & 49.92 & 45.86 \\
        \bottomrule
        \end{tabular}}
        \label{tab:language_only}
\end{table}

\section{Limitations and future directions}
\label{sec:limitations}
Despite the promising results of \model{}, our work is the first study in this direction and some limitations are yet to be addressed. 
%The first shortcoming concerns the representation capabilities of the \nftovec{} encoder, which uses one single token to encode the NeRF input. 
%Following the approach of existing MLLMs like PointLLM~\cite{pointllm}, GPT4Point~\cite{gpt4point}, and LLaVA~\cite{llava}, it may be worth exploring the development of richer representations using multiple tokens.
The first limitation is that \nftovec{} can only process MLPs, which restricts our model to MLP-only NeRFs. However, thanks to the rapid advancements in meta-networks, it may become very soon possible to extend \model{} to more complex \nerf{} architectures, such as InstantNGP~\cite{instant}. For instance, the approach by~\cite{lim2024graph} suggests the feasibility of processing various input architectures, although it is currently limited to small networks. 
The second shortcoming is that our framework has been tested solely on object-centric \nerf{}s. Expanding its application to \nerf{}s representing entire scenes would be a compelling direction for future research.

\section{Concluding remarks}
\label{sec:remarks}
This paper addressed the novel task of creating a language assistant for \nerf{}. We have tackled this problem by leveraging recent advances in MLLMs and meta-networks processing neural fields.  We have shown that it is feasible and effective to directly process the weights of a NeRF to project it into the input embedding space of an LLM. Building on our previous work~\cite{amaduzzi2024llana}, we have presented ObjaNeRF--Text, a benchmark for NeRF-language understanding that includes 280K annotated NeRFs with text-based conversations. Furthermore, we have scaled \model{} to a larger LLM and conducted a detailed analysis of the impact of model size on NeRF-language performance. Finally, we have extended such analysis to existing MLLMs, offering insights into the scalability and effectiveness of different architectures.
\section*{Acknowledgements}
%We acknowledge the CINECA award under the ISCRA initiative, for the availability of high-performance computing resources and support.
We acknowledge ISCRA for awarding this project access to the LEONARDO supercomputer, owned by the EuroHPC Joint Undertaking, hosted by CINECA (Italy). 

This work was partially funded by ``FSE+ 2021-2027 ai sensi dell’art. 24, comma 3, lett. a), della Legge 240/2010 e s.m.i. e del D.G.R. 693/2023  (RIF. PA: 2023-20090/RER - CUP: J19J23000730002)".

\bibliographystyle{IEEEtran}
\bibliography{main}

%\vspace{-1cm}
\begin{IEEEbiography}
[{\includegraphics[width=1in,height=1.25in,clip,keepaspectratio]{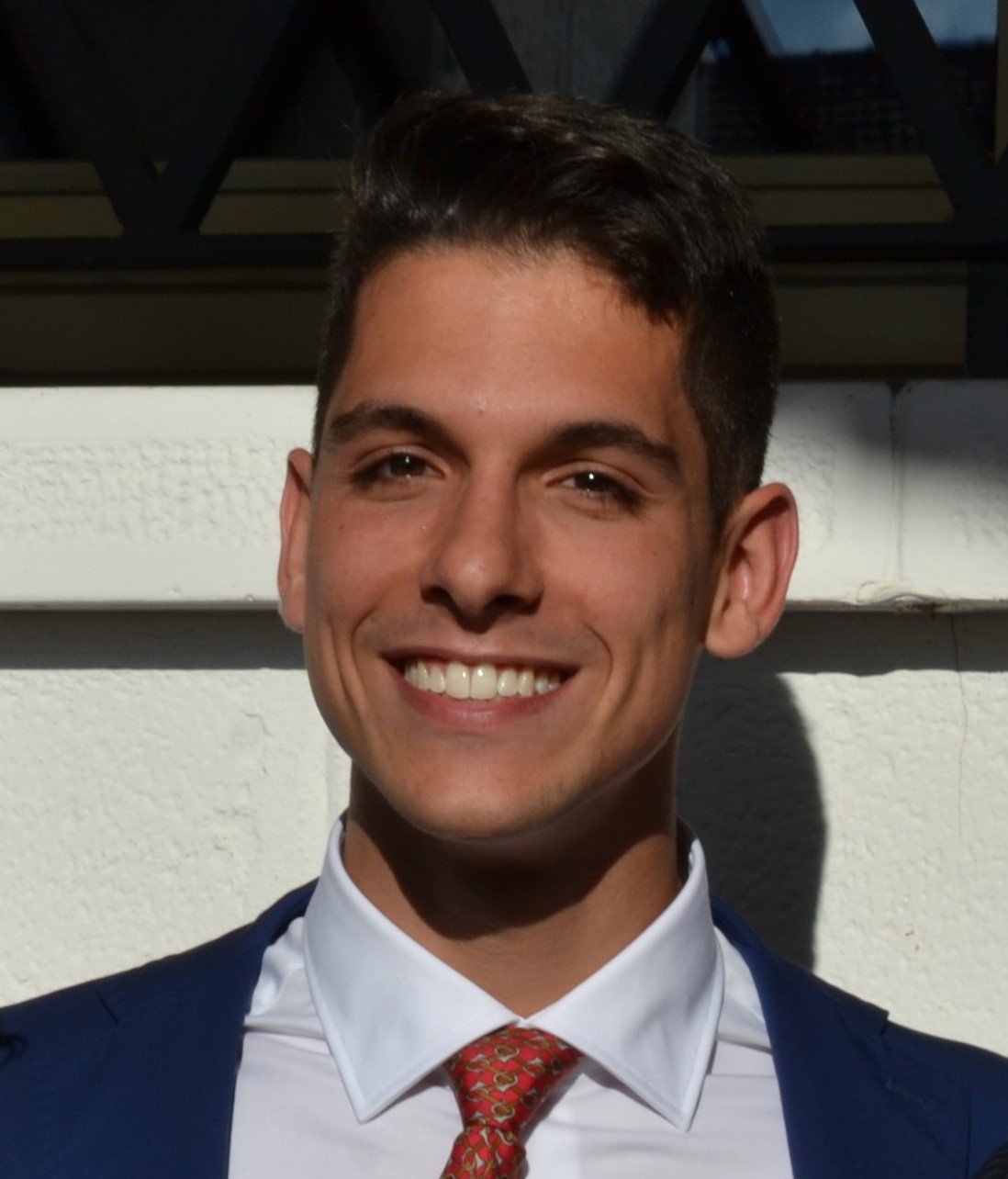}}]
{Andrea Amaduzzi} is a fourth-year PhD student at the Computer Vision Laboratory (CVLAB), University of Bologna. Prior to his doctoral studies, he worked as a Computer Vision Software Engineer at Datalogic. He authored several research papers covering a range of subjects, including 3D computer vision and multimodal learning.
\end{IEEEbiography}
%\vspace{-0.5cm}
\begin{IEEEbiography}
[{\includegraphics[width=1in,height=1.25in,clip,keepaspectratio]{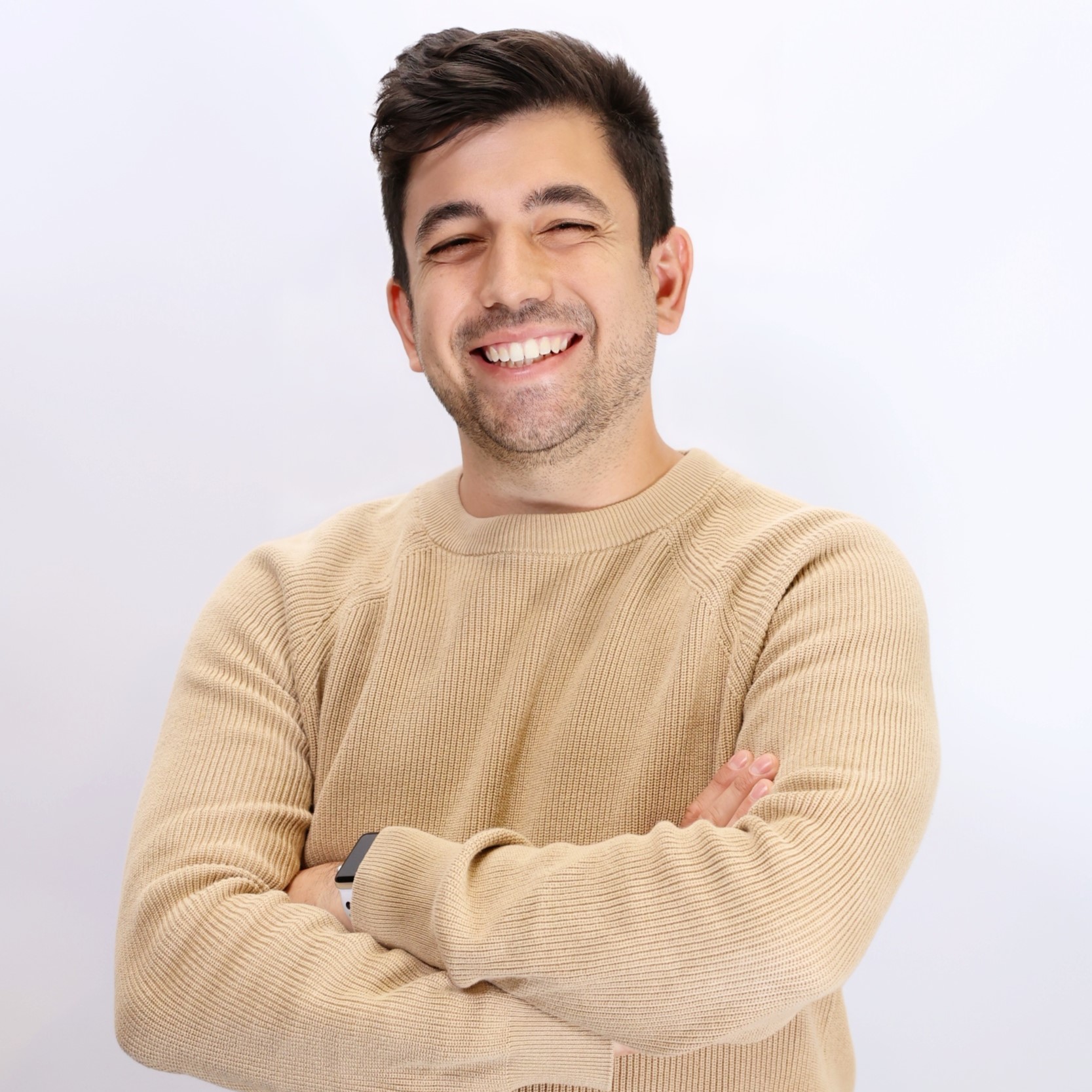}}]
{Pierluigi Zama Ramirez} received his PhD in Computer Science and Engineering in 2021. He has been a Research Intern at Google for 6 months and is currently a Post-Doc at the University of Bologna. He co-authored more than 20 publications on computer vision research topics such as semantic segmentation, depth estimation, optical flow, domain adaptation, virtual reality, and 3D computer vision.
\end{IEEEbiography}
%\vspace{-0.5cm}
\begin{IEEEbiography}
[{\includegraphics[width=1in,height=1.25in,clip,keepaspectratio]{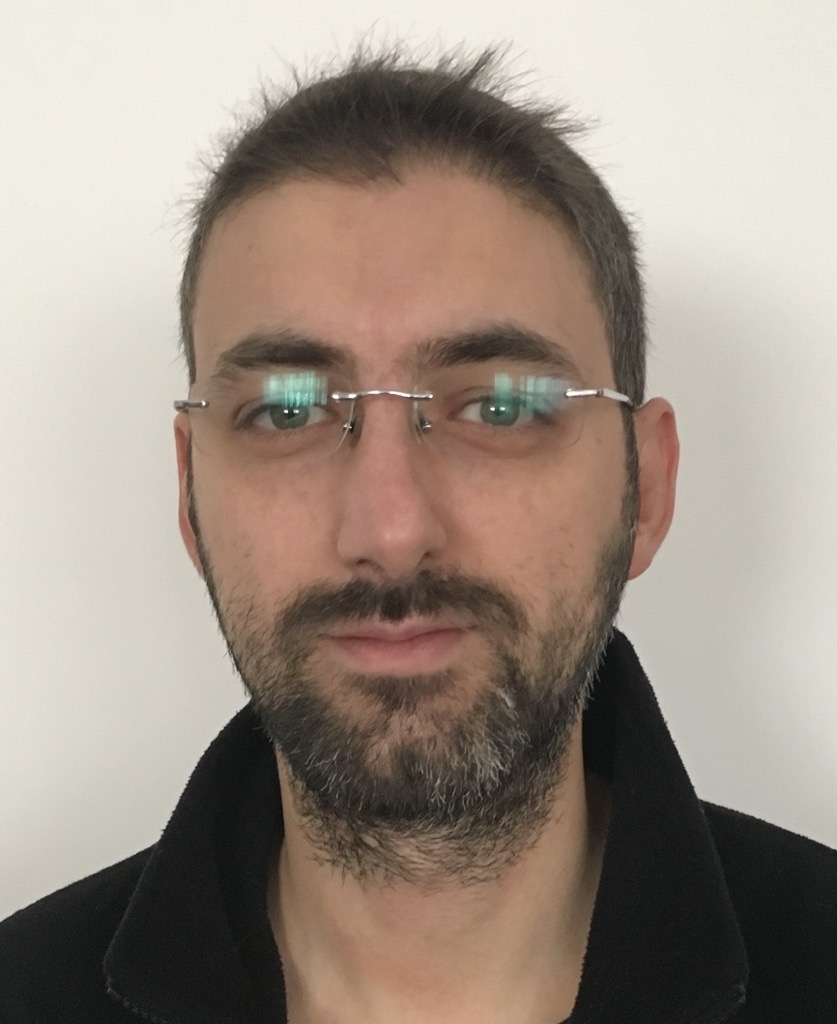}}]{Giuseppe Lisanti} is currently an Associate Professor in the Department of Computer Science and Engineering at the University of Bologna. He has co-authored over 50 publications, with his primary research interests focus on computer vision and the application of deep learning to computer vision problems.
He actively collaborates with other research centres and has participated in various roles in multiple research projects. In 2017, he received the Best Paper Award from the IEEE Computer Society Workshop on Biometrics.
\end{IEEEbiography}
%\vspace{-0.5cm}
\begin{IEEEbiography}
[{\includegraphics[width=1in,height=1.25in,clip,keepaspectratio]{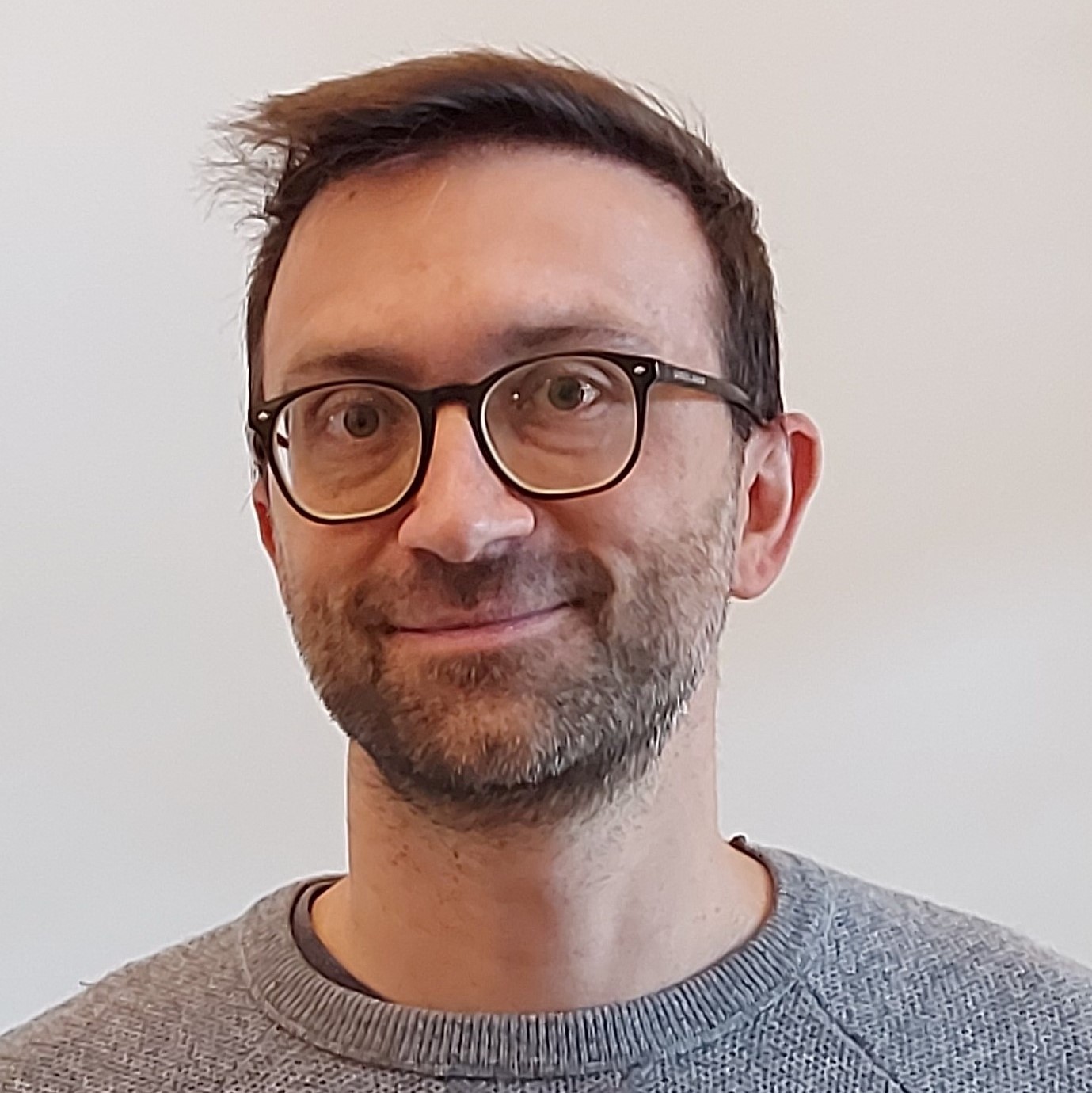}}]{Samuele Salti} is currently an associate professor at the Department of Computer Science and Engineering (DISI) of the University of Bologna, Italy.  His main research interest is computer vision, mainly 3D computer vision and machine/deep learning applied to computer vision problems.
Dr. Salti has co-authored more than 60 publications and 8 international patents. In 2020, he co-founded the start-up eyecan.ai. 
\end{IEEEbiography}
%\vspace{-0.5cm}
\begin{IEEEbiography}
[{\includegraphics[width=1in,height=1.25in,clip,keepaspectratio]{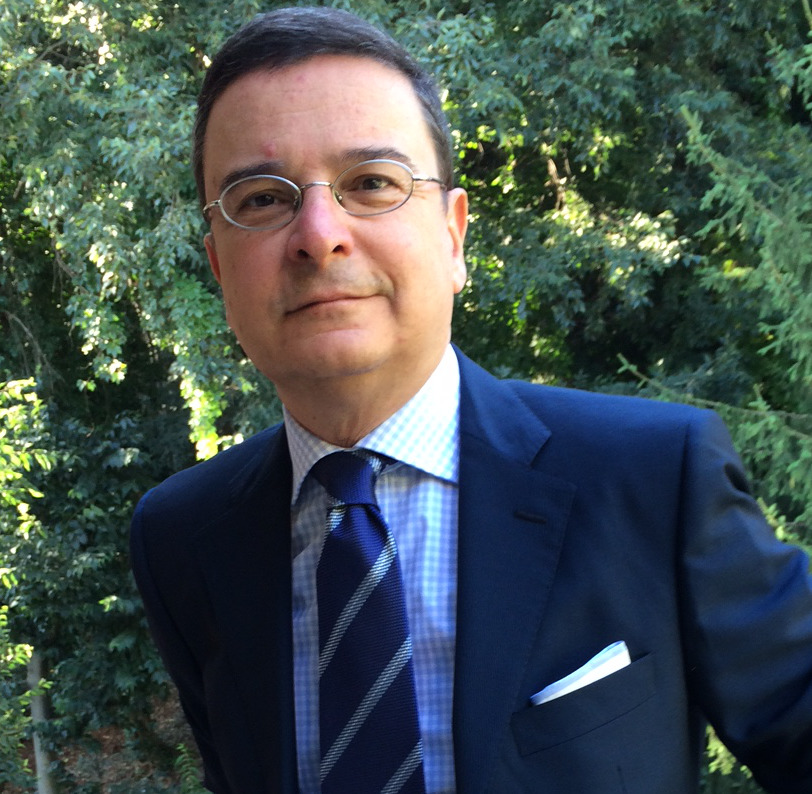}}]
{Luigi Di Stefano}
received a PhD degree in electronic engineering and computer science from the University of Bologna in 1994. He is a full professor at the Department of Computer Science and Engineering, University of Bologna, where he founded and led the Computer Vision Laboratory (CVLab). His research interests include image processing, computer vision, and machine/deep learning. He is the author of more than 150 papers and several patents. He has been a scientific consultant for major computer vision and machine learning companies.  He is a member of the IEEE Computer Society and the IAPR-IC.
\end{IEEEbiography}

\end{document}